\documentclass{article}
\pdfoutput=1

\usepackage[latin9]{inputenc}
\usepackage{units}
\usepackage{amsmath}
\usepackage{amssymb}

\usepackage{graphicx} 
\usepackage{subfigure}

\usepackage{natbib,natbibspacing}

\usepackage{algorithm}
\usepackage{algorithmic}

\usepackage{nips12submit_e,times}
\usepackage{multirow}

\usepackage{amsfonts}
\usepackage{algorithmic}\usepackage{algorithm}

\newcommand{\bb}[1]{\boldsymbol{\mathrm{#1}}}

\def\Tr{\mathrm{T}}

\newcommand{\RR}{\mathbb{R}}

\newcommand{\vv}{\bb{v}}

\newcommand{\xx}{\bb{x}}

\newcommand{\Rr}{\bb{R}}

\newcommand{\Xx}{\bb{X}}

\newcommand{\Vv}{\bb{V}}
\newcommand{\Uu}{\bb{U}}
\newcommand{\Dd}{\bb{D}}
\newcommand{\Ii}{\bb{I}}

\newcommand{\Ww}{\bb{W}}
\newcommand{\Ll}{\bb{L}}

\newcommand{\Llambda}{\bb{\Lambda}}

\newcommand{\tr}{\mathrm{tr}\,}

\title{Multimodal diffusion geometry by joint diagonalization of Laplacians}

\author{
Davide Eynard \\
Institute of Computational Science\\
Faculty of Informatics\\
Universit\`a della Svizzera italiana\\
Via G. Buffi 13, Lugano 6904, Switzerland\\
\small{\texttt{davide.eynard@usi.ch}} \\
\And
Klaus Glashoff \\
Department of Mathematics \\
University of Hamburg \\
Hamburg, Germany\\
\small{\texttt{klaus.glashoff@math.uni-hamburg.de}} \\
\AND
Michael M. Bronstein \\
Institute of Computational Science\\
Faculty of Informatics\\
Universit\`a della Svizzera italiana\\
Via G. Buffi 13, Lugano 6904, Switzerland\\
\small{\texttt{michael.bronstein@usi.ch}} \\
\And
Alexander M. Bronstein \\
School of Electrical Engineering\\
Faculty of Engineering\\
Tel Aviv University\\
Ramat Aviv 69978, Israel\\
\small{\texttt{bron@eng.tau.ac.il}} \\
}

\begin{document}

\maketitle

\begin{abstract}


We construct an extension of diffusion geometry to multiple modalities through joint approximate diagonalization of Laplacian matrices.
This naturally extends classical data analysis tools based on spectral geometry, such as diffusion maps and spectral clustering. 
We provide several synthetic and real examples of manifold learning, retrieval, and clustering demonstrating that the joint diffusion geometry frequently better captures the inherent structure of multi-modal data.
We also show that many previous attempts to construct multimodal spectral clustering can be seen as particular cases of joint approximate diagonalization of the Laplacians.

\end{abstract}

\section{Introduction}

The Laplacian operator and related constructions play a pivotal role
in a wide range of applications in machine learning, pattern recognition,
and computer vision community. It has been shown that many problems
in these fields boil down to finding some eigenvectors and eigenvalues
of a Laplacian constructed on some high-dimensional data.
Important examples include {\em spectral clustering} (\citet{Ng01onspectral}) where clusters
are determined by the first eigenvectors of the Laplacian; {\em eigenmaps} (\citet{Belkin02laplacianeigenmaps}) 
and more generally {\em diffusion maps} (\cite{Coifman}), where one tries to find
a low-dimensional manifold structure using the first smallest eigenvectors
of the Laplacian; and {\em diffusion metrics} (\cite{Coifman05geometricdiffusions}) measuring the ``connectivity''
of points on a manifold and expressed through the eigenvalues and
eigenvectors of the Laplacian. 
Other applications heavily relying
on the properties of the Laplacian include 
{\em spectral
graph partitioning} (\cite{DingHeetal2001}), {\em spectral hashing} (\cite{weiss2008spectral}), spectral correspondence, image segmentation (\cite{Shi97normalizedcuts}), 
and 
shape analysis (\cite{Levy:2006:LET:1136647.1136965}). Because of
the intimate relation between the Laplacian operator, Riemannian geometry,
and diffusion processes, it is common to encounter the
umbrella term {\em spectral} or {\em diffusion geometry} in relation to the above
problems.

These applications have been considered mostly 
in the context of uni-modal data, i.e., a single data space. However,
many applications involve observations and measurements of data done
using different modalities, such as multimedia documents  (\cite{weston2010large,rasiwasia2010new,mcfee2011learning}), audio and video (\cite{kidron2005pixels,alameda2011finding}), or medical imaging modalities like PET and CT (\cite{bronstein2010data}). 
Such problems of multimodal (or multi-view)
data analysis have gained increasing interest in the computer vision
and pattern recognition communities, however, there have been only
few attempts extending the powerful spectral methods to such settings.


In this paper, we propose a general framework allowing to extend different
diffusion and spectral methods to the multimodal setting by finding a common eigenbasis of multiple Laplacians. Numerically, this problem is posed as {\em approximate
joint diagonalization} of several matrices. Such methods have received limited attention in the numerical mathematics community (\cite{Bunse-Gerstner:1993}) and have been employed for joint diagonalization of covariance matrices in blind source separation applications by \cite{CardosoBlind1993,Cardoso96jacobiangles,YeredorJD202,ZieglerBlindsource05}. To the best of our knowledge, this is the first time they are applied to spectral embeddings. 
Besides providing a principled approach to data fusion, our framework gives a theoretical explanation to existing methods for multimodal data analysis. In particular, we show that many recent works on multi-view clustering by \cite{de2005spectral,MaAnchor2008,tang2009clustering,Cai2011,kumarco} can be considered a particular instance of our framework.

\section{Background}

Let us be given some data represented as a $k$-dimensional manifold $X \subset \RR^d$, embedded into a $d$-dimensional Euclidean space. In many applications $d$ is very large while the intrinsic dimension of the data $k$ is small, and one tries to study the structure of the manifold rather than its $d$-dimensional embedding. 
Such a structure can be characterized by the means of the {\em Laplace-Beltrami operator}.  
In the discrete setting, the manifold is often represented by a weighted graph with vertices $\{\xx_{1},\hdots,\xx_{n}\} \subset X$ and edge weights $w_{ij}=k(\xx_{i},\xx_{j})$ representing local connectivity using e.g. Gaussian kernel (see \cite{Luxburg}). 
The Laplace-Beltrami operator can be discretized\footnote{There exist many different constructions of the discrete Laplacian. For the sake of simplicity, we adopt the symmetric Laplacian. Our framework is applicable to other discretization as well.} as $\Ll=\Dd^{-1/2}(\Dd - \Ww)\Dd^{-1/2}$, where $\Ww = (w_{ij})$ and $\Dd = \mathrm{diag}(\sum_{j}w_{ij})$. Such a discretization is often referred to as {\em symmetric normalized Laplacian} and admits a unitary diagonalization $\Ll=\Vv\Llambda\Vv^{\Tr}$, $\Vv\Vv^{\Tr}=\Ii_{n}$
with the eigenvalues $\lambda_{1}=0\leq\lambda_{2}\leq\hdots\leq\lambda_{n}$.
Geometric constructions associated with eigenvectors and eigenvalues of the Laplacian play an important role in machine learning, since several archetypical  problems can be formulated in these terms:

%
%
{\bf Eigenmaps. } 
Non-linear dimensionality reduction methods try to capture the intrinsic low-dimensional structure of the manifold $X$. 
Belkin and Niyogi (\citeyear{Belkin02laplacianeigenmaps}) showed that finding
a neighborhood-preserving $k$-dimensional embedding of $X$ can be posed as the 
minimum eigenvalue problem, 
\begin{eqnarray}
\min_{\Vv\in\RR^{n\times k}}\tr(\Vv^{\Tr}\Ll \Vv) \,\,\, \mathrm{s.t.} \,\,\, \Vv^{\Tr}\Vv=\Ii. \label{eq:belkin}
\end{eqnarray} 
This problem is minimized by setting $\Vv$ to be the matrix containing the first $k$ eigenvectors of $\Ll$, thus effectively embedding the data by means of the eigenfunctions of the Laplace-Beltrami operator (the null eigenvector is usually discarded). Such an embedding is referred to as {\em Laplacian eigenmap}. 
More generally, a {\em diffusion map} is given as a mapping of the form $\Psi = (K(\lambda_2)\vv_2,\hdots, K(\lambda_k) \vv_{k})$, 
where $K(\lambda)$ is some transfer function acting as a ``low-pass filter'' on eigenvalues $\lambda$ (\cite{Coifman05geometricdiffusions, Coifman}).

{\bf Diffusion distances. } 
Coifman et al. (\citeyear{Coifman05geometricdiffusions, Coifman}) related the eigenmaps to heat diffusion and random processes on manifolds and defined a family of {\em diffusion metrics} that in the most general setting can be written as 
\begin{equation}
d^{2}(\xx_{i},\xx_{j})=\sum_{l} K(\lambda_l) (v_{il} - v_{jl})^2 = \| \Psi(\xx_i) - \Psi(\xx_j) \|^2_2. 
\label{eq:diffdist}
\end{equation}
Particular choice of $K(\lambda) = e^{-\lambda t}$ gives the {\em heat diffusion distance}, related to the connectivity of points $\xx_i, \xx_j$ on the manifold by means of diffusion process of length $t$. 
Such distances are intrinsic and thus invariant to manifold embedding and are robust to topological noise.

 {\bf Spectral clustering. } 
\cite{Ng01onspectral} showed a very efficient and robust clustering approach based on the observation that the multiplicity of the null  eigenvalue of $\Ll$ is equal to the number of connected components of $X$. The corresponding eigenvectors act as indicator functions of these components. 
  Embedding the data using these eigenvectors and then applying some standard clustering algorithm such as K-means was shown to produce significantly better results than clustering the high-dimensional data directly.

\section{Multimodal diffusion geometry}

Recently, we witness increasing popularity of attempts to analyze different ``views'' or modalities of data. 
Such data can be modeled as 
$m$ different manifolds $X^1 \subset \RR^{d_1}, \hdots, X^m \subset \RR^{d_m}$, which can have embeddings of different dimensionality ($d_1,\hdots, d_m$) and sometimes different structure.
%
%
We are interested in analyzing these manifolds simultaneously in order to extract their joint intrinsic structure. 
We assume that we are given $n$ corresponding samples $\{(\xx^i_1,\hdots,\xx^i_n)\subset \RR^{d_i}\}_{i=1}^m$ on the manifolds 
and can construct the Laplacian matrices $\Ll_1,\hdots, \Ll_m$ as described in the previous section.

\begin{figure}[t!]
\centering
%
%
\includegraphics[width=1\linewidth]{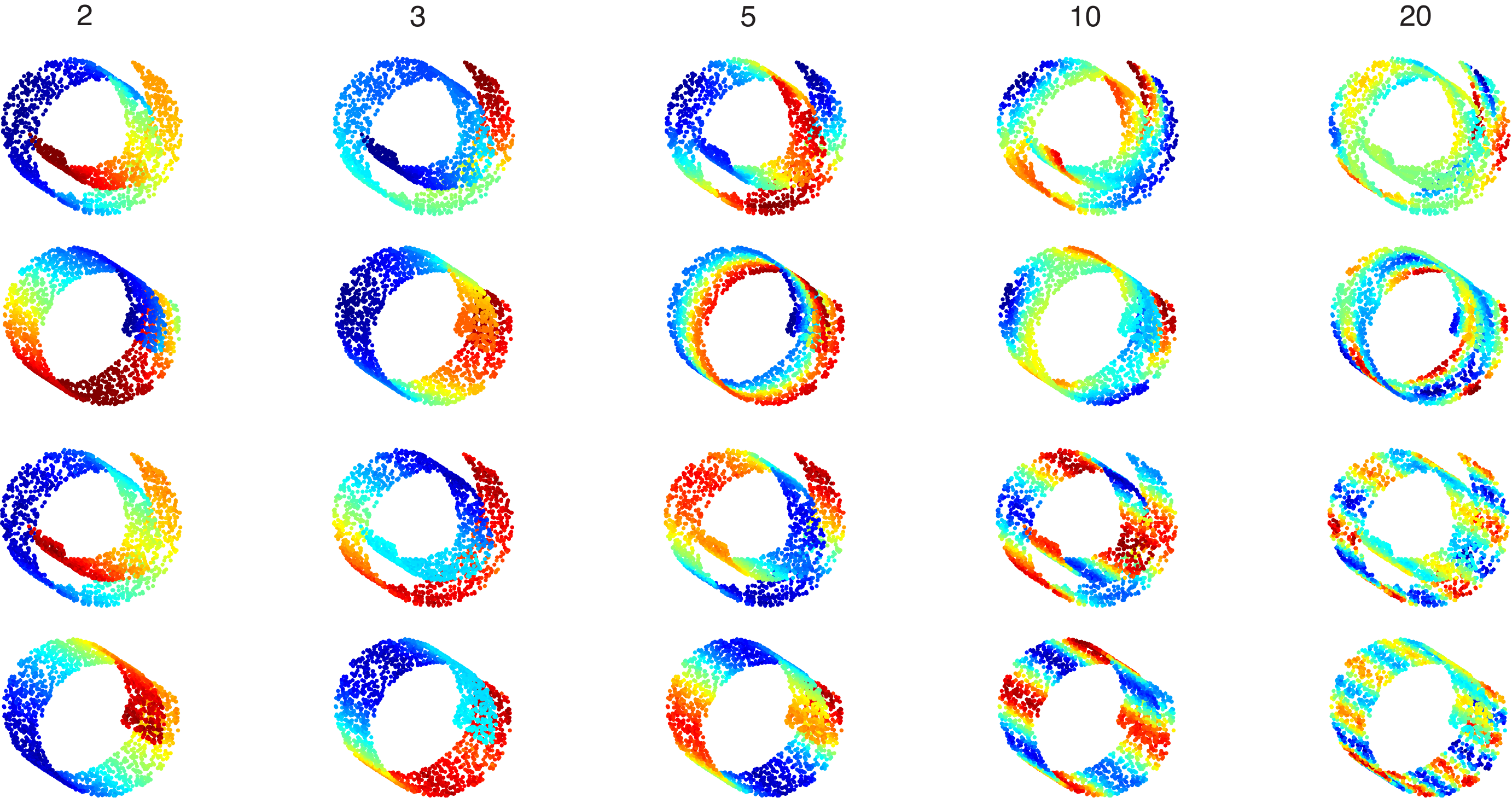} 
%
%
%
\caption{\label{fig:evecs}First and second rows: eigenfunctions of the Laplacians of two modalities of the Swiss roll. 
Third and fourth rows: joint eigenfunctions of the two Laplacians computed using JADE. Hot colors represent positive values; cold colors represent negative values. }
\end{figure}

Trying to use the eigenvectors $\Vv_1,\hdots,\Vv_m$ of the Laplacian matrices $\Ll_1,\hdots, \Ll_m$ is problematic: 
for a set of eigenvectors corresponding to an eigenvalue with multiplicity greater than one, we can talk only of eigen sub-space, and any basis spanning it is a valid set of eigenvectors. 
%
As a result, the eigenvectors of the Laplacians in different modalities can be substantially different (Figure~\ref{fig:evecs}, top).

{\bf Joint diagonalization. } A solution is to try to find the eigenbasis of the Laplacians {\em simultaneously}. 
This problem is known as {\em joint diagonalization} and consists of finding a set of joint orthogonal eigenvectors $\bar{\Vv}$ 
such that $\bar{\Vv}^\Tr \Ll_i \bar{\Vv} = \Llambda_i$ are diagonal matrices of the eigenvalues of $\Ll_i$. Such a common eigenbasis solves the inherent ambiguity in the definition of the eigenvectors and ``couples'' different modalities (Figure~\ref{fig:evecs}, bottom). 
However, due to differences between the modalities and the presence of noise, the Laplacian matrices $\Ll_1,\hdots, \Ll_m$ rarely have a joint eigenbasis (iff they commute). 
It is still possible to find an {\em approximate} joint diagonalization by solving 
\begin{equation}
\min_{\bar{\Vv}}\sum_{i=1}^{m} \mathrm{off}(\bar{\Vv}^{\Tr}\Ll_{i}\bar{\Vv}), \,\,\, \mathrm{s.t.} \,\,\, \bar{\Vv}^\Tr \bar{\Vv} = \Ii, \label{eq:costfunction}
\end{equation}
where $\mathrm{off}(\Xx)$ is some off-diagonality criterion, e.g. the sum of squared off-diagonal elements,  $\mathrm{off}(\Xx) = \| \Xx - \mathrm{diag}(\Xx)\|_\mathrm{F}^{2}$.  
In this case, $\bar{\Vv}^\Tr \Ll_i \bar{\Vv}$ are only approximately diagonal; we refer to the average of the diagonal elements $\bar{\Llambda} = \frac{1}{m}\sum_{i=1}^m \mathrm{diag}(\bar{\Vv}^\Tr \Ll_i \bar{\Vv})$ as the {\em joint approximate eigenvalues} of $\Ll_1,\hdots, \Ll_m$. 
This definition allows us to naturally extend the diffusion geometric methods discussed in the previous section (eigenmaps, diffusion distances, spectral clustering, etc.)
to the multimodal setting by simply replacing the eigenvalues and eigenvectors of a single Laplacian $\Ll_i$ by the joint eigenvectors  $\bar{\Vv}$ and eigenvalues $\bar{\Llambda}$ of multiple Laplacians $\Ll_1, \hdots, \Ll_m$.

{\bf Numerical computation. } 
A numerical method for joint diagonalization based on a modified  \emph{Jacobi iteration} traces back to \cite{Bunse-Gerstner:1993}, and it has been used at about the same time by Cardoso and Souloumiac (\citeyear{CardosoBlind1993,Cardoso96jacobiangles}) for joint diagonalization of covariance matrices in the context of blind source separation. 
%
%
The idea of the standard Jacobi method for eigenvalue calculation is to apply a sequence of plane rotations in order to sequentially minimize the off-diagonal elements of the given matrix. The rotation is applied ``in-place'' and does not require matrix multiplication. 
In the modified Jacobi method (referred to as JADE), the rotations are applied to reduce the off-diagonality criterion (\ref{eq:costfunction}) in each step. Let $\Rr_{pqcs}$ the (complex) rotation matrix the entries of which are equal to those of the identity matrix except for the elements 
\begin{equation}
\left(\begin{array}{cc}
r_{pp} & r_{pq}\\
r_{qp} & r_{qq}
\end{array}\right)=\left(\begin{array}{cc}
c & \bar{s}\\
-s & \bar{c}
\end{array}\right)
\end{equation}
where $\mid c\mid^{2}+\mid s\mid^{2}=1$. \citet{Cardoso96jacobiangles}
show that the problem 
\begin{equation}
\min_{\mid c\mid^{2}+\mid s\mid^{2}=1}\sum_{i=1}^{m} \mathrm{off}(\Rr_{pqcs}^{\Tr}\Ll_{i}\Rr_{pqcs}) \,\,\, 
\end{equation}
has a simple explicit solution based on a $3\times3$ eigenvalue problem. 
JADE is one of the most common algorithms in the field of joint diagonalization and has complexity comparable to that of the standard Jacobi method.  There are other algorithms, like the  ACDC method of \citet{YeredorJD202}, as well as different versions of the idea of minimizing a suitable cost function on the Stiefel manifold (\citet{Rahbar00geometricoptimization}).

{\bf Analytic computation. }
In the spectral clustering problem, we are looking for the null eigenvectors of the Laplacian. 
Assuming that the first $k$ eigenvalues of the Laplacians are zero, we want to find $\bar{\Vv} \in \RR^{n\times k}$ such that $\Ll_i \bar{\Vv} = 0$ for all $i=1,\hdots, m$ and $\bar{\Vv}^\Tr \bar{\Vv} = \Ii$  by reformulating~(\ref{eq:costfunction}) as  
\begin{equation}
\min_{\bar{\Vv} \in \RR^{n\times k}}  \sum_{i=1}^m \| \Ll_i \bar{\Vv} \|^2_\mathrm{F}, \,\,\, \mathrm{s.t.} \,\,\, \bar{\Vv}^\Tr \bar{\Vv} = \Ii. 
\label{eq:analytic1}
\end{equation}
Since $\sum_{i=1}^m\| \Ll_i \bar{\Vv} \|^2_\mathrm{F}= \tr (\bar{\Vv}^\Tr(\sum_{i=1}^m \Ll_i^\Tr \Ll_i )\bar{\Vv})$, the problem can be equivalently recast as single-modality clustering with the ``average''  Laplacian matrix 
$\bar{\Ll} =  \sum_{i=1}^m \Ll_i^\Tr \Ll_i$. 
%
%
%
%
%
%
We can also consider other averaging operators, e.g. weighted arithmetic mean 
$\bar{\Ll} =  \sum_{i=1}^m w_i\Ll_i$ or harmonic mean 
$\bar{\Ll} =  (\sum_{i=1}^m \Ll^{-1}_i)^{-1}$. 
We discuss these methods in the next section. 

%
For zero eigenvalues, (\ref{eq:analytic1}) is akin to (\ref{eq:costfunction}), which justifies the successful use of such ``averaging'' methods in problems of multimodal spectral clustering (\cite{MaAnchor2008,Cai2011}).   
However, iterative methods such as JADE explicitly minimizing the off-diagonality criterion (\ref{eq:costfunction}) are more generic and applicable to settings where one has to find all or many joint eigenvectors, e.g., for computing eigenmaps or diffusion distances.

%
%

\section{Relation to previous works}

There have been numerous recent works on multimodal spectral-type clustering proposing different ways of fusing multiple modalities based on different principles. 
Considering these methods through the prism of joint diagonalization, we show many commonalities and equivalences between algorithms stemming from different motivations and coming from various communities.
%
%
%
\cite{MaAnchor2008} considered detection of shots in video sequences using fusion of video and audio information, employing for this purpose spectral clustering of a Laplacian created as a weighted arithmetic mean of each modality Laplacian. 
\cite{tang2009clustering} 
used low-rank factorization of the weight matrix, trying to find a common factor $\Uu$ such that $\Ww_i \approx \Uu \Llambda_i \Uu^\Tr$ 
by solving 
\begin{equation}
\min_{\Uu \in \RR^{n\times k}, \Lambda_i \in \RR^{n\times n}} \sum_{i=1}^m \| \Ww_i - \Uu \Llambda_i \Uu^\Tr \|_\mathrm{F}^2, 
\label{eq:tang1}
\end{equation}
%
using the quasi-Newton method. 
Besides the fact that the factorization is applied to the weight matrix (it can be equivalently applied to the Laplacian), we see here a (non-orthogonal) joint diagonalization problem with an off-diagonality criterion considered by \cite{YeredorJD202}.

\cite{Cai2011} proposed a method for {\em multiview spectral clustering} (MVSC) by solving\footnote{\cite{Cai2011} also impose a non-negativity constraint on the matrix $\Vv$ in order to obtain cluster indicators directly and bypass the K-means clustering stage. We ignore this additional constraint for the simplicity of discussion; such a constraint can be added to all the problems discussed in this paper. }
\begin{equation}
\min_{\Vv_i, \Vv \in \RR^{n\times k} } \sum_{i=1}^{m}\tr(\Vv_{i}^\Tr\Ll_{i}\Vv_{i})+\alpha \|\Vv_{i}-\Vv\|_\mathrm{F}^{2}\,\,\, \mathrm{s.t.} \,\,\, \Vv^\Tr \Vv = \Ii
\label{eq:Caiminimization}
\end{equation}
The authors show that this problem can be equivalently posed as 
\begin{equation}
\max_{\Vv \in \RR^{n\times k} } \tr \left( \Vv^\Tr  \textstyle \sum_{i=1}^{m}\left(\Ll_{i} + \alpha \Ii \right)^{-1} \Vv \right)\,\,\, \mathrm{s.t.} \,\,\, \Vv^\Tr \Vv = \Ii, 
\label{eq:Caiminimization1}
\end{equation}
and then employ an iterative algorithm to find the solution $\Vv$. 
First, we observe that problem (\ref{eq:Caiminimization}) consists of $m$ minimum-eigenvalue problems w.r.t. bases $\Vv_i$, with the addition of a coupling term, 
encouraging $\Vv_i$ as close as possible to some common basis $\Vv$ (note that the authors do not impose orthogonality constraints $\Vv_i^\Tr \Vv = \Ii$, but for $\alpha \gg 0$, the proximity to orthogonal $\Vv$ makes $\Vv_i$ approximately orthogonal).
Thus, it is possible to interpret~(\ref{eq:Caiminimization})  as a kind of joint diagonalization criterion. 
Second, problem (\ref{eq:Caiminimization1}) can be rewritten as a minimum eigenvalue problem 
\begin{equation}
\min_{\Vv \in \RR^{n\times k} } \tr \left( \Vv^\Tr  \left( \textstyle \sum_{i=1}^{m}\left(\Ll_{i} + \alpha \Ii \right)^{-1} \right)^{-1} \Vv \right)\,\,\, \mathrm{s.t.} \,\,\, \Vv^\Tr \Vv = \Ii, 
\label{eq:Caiminimization2}
\end{equation}
whose solution is given by the matrix composed of the first $k$ eigenvectors of the matrix 
$\left( \textstyle \sum_{i=1}^{m}\left(\Ll_{i} + \alpha \Ii \right)^{-1} \right)^{-1}$. For $\alpha>0,$ this a regularized version of the \emph{harmonic mean} of the Laplacian matrices. We can thus regard the method of \cite{Cai2011} as a particular instance of our joint diagonalization approach discussed in the previous section. 
%

\cite{kumarco} proposed 
the {\em centroid co-regularization} approach for multimodal clustering based on the minimization of  
\begin{equation}
\min_{\Vv, \Vv_i \in \RR^{n\times k}  } \sum_{i=1}^m \tr ( \Vv_i^\Tr  \Ll_i \Vv_i ) - \alpha  \tr ( \Vv_i \Vv_i^\Tr \Vv \Vv^\Tr ) \,\,\, \mathrm{s.t.} \,\,\, \Vv_i^\Tr \Vv_i = \Ii; \,\,\, \Vv^\Tr \Vv = \Ii. 
\label{eq:kumar2}
\end{equation}
%
This function is alternatingly minimized, first with respect to the $\Vv_i$, then with respect to $\Vv$. 
Problems~(\ref{eq:kumar2}) and~(\ref{eq:Caiminimization}) are similar in their spirit (the first one uses dissimilarity $\| \Vv_i - \Vv \|_\mathrm{F}^2$ as coupling term, while the second one the similarity $\tr ( \Vv_i \Vv_i^\Tr \Vv \Vv^\Tr ) = \| \Vv_i^\Tr \Vv \|_\mathrm{F}^2$)), and  fall under our joint diagonalization framework.

We must stress that these methods were developed for clustering problems where one has to find the null eigenvectors, and do not adapt easily to other applications of diffusion geometry where one has to find many or all joint eigenvectors of the Laplacians (e.g., computation of diffusion distances).  
In particular, iterative solvers used in \cite{tang2009clustering,kumarco,Cai2011} do not scale up to such cases. 
On the other hand, algorithms such as modified Jacobi iteration (JADE) are made for finding a full set of joint eigenvectors and have the complexity akin to standard Jacobi iteration. Further speed-up might be achieved by making explicit use of the sparse structure of the Laplacian matrices, which is not taken advantage of in JADE.

\begin{figure}[t!]
\centering
\begin{minipage}[b]{0.3\linewidth}
\includegraphics[width=1\linewidth,height=0.7\linewidth]{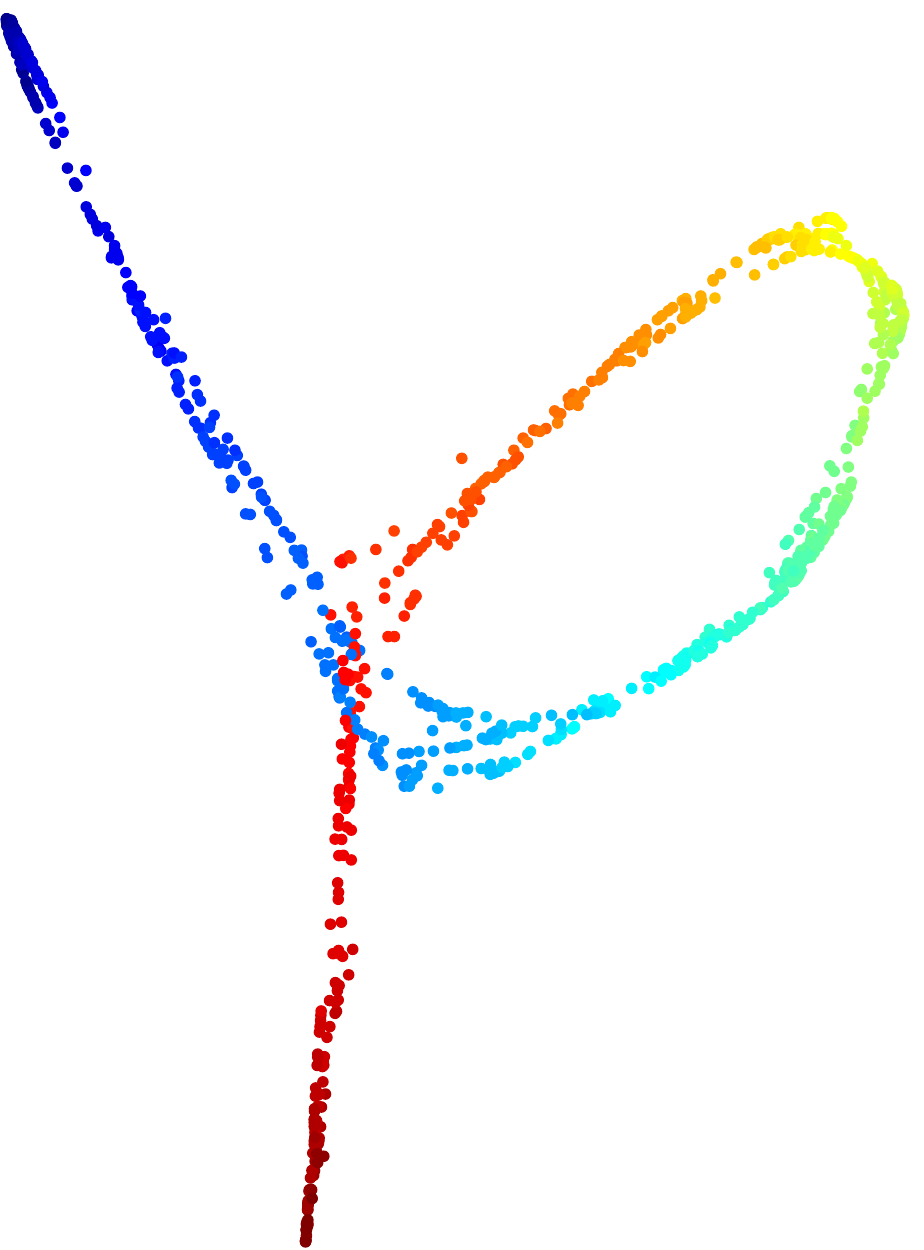} 
\end{minipage}\hspace{5mm}
\begin{minipage}[b]{0.3\linewidth}
\includegraphics[width=1\linewidth,height=0.7\linewidth]{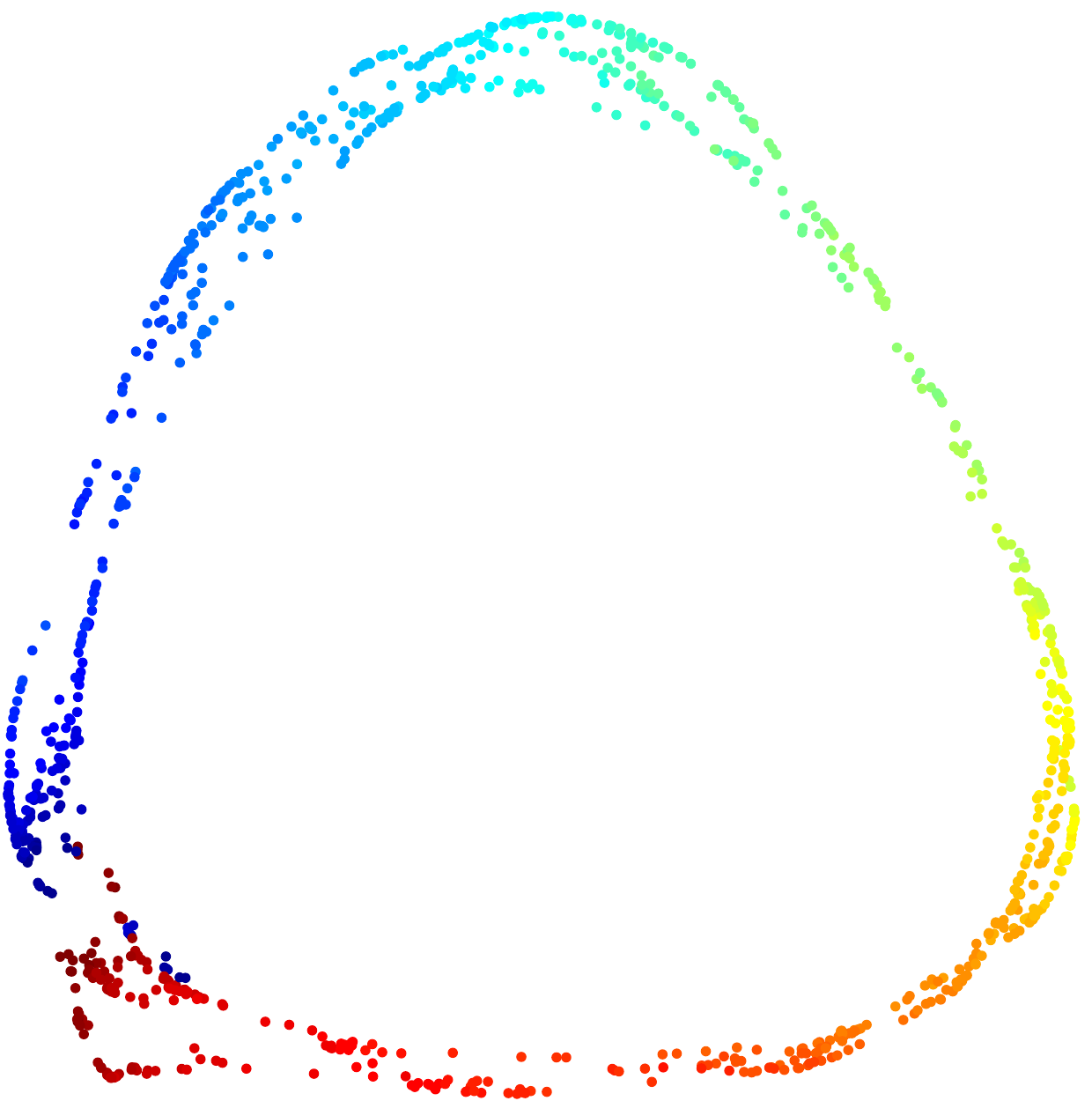} 
\end{minipage}\hspace{5mm}
\begin{minipage}[b]{0.3\linewidth}
\includegraphics[width=1\linewidth,height=0.7\linewidth]{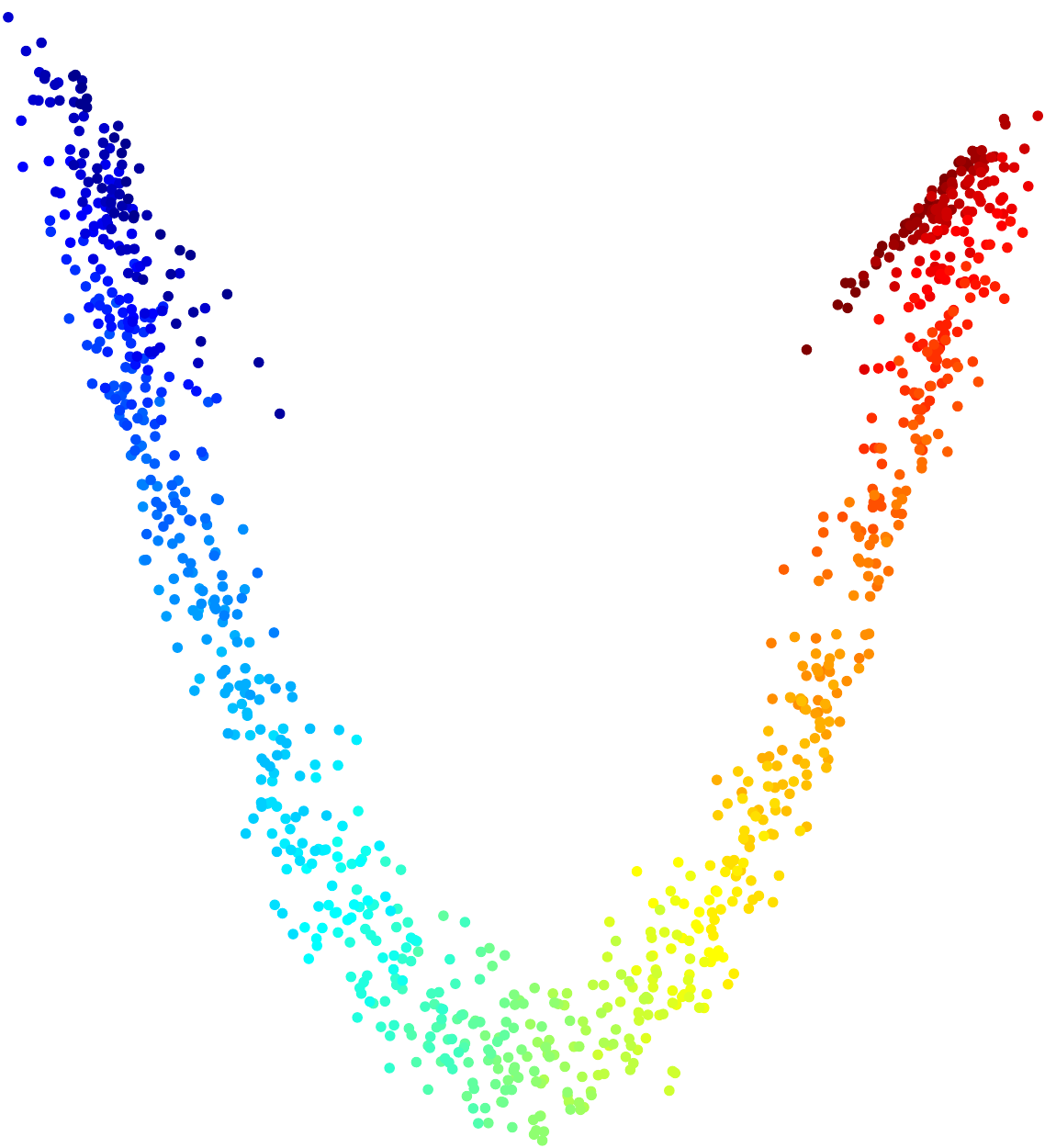} 
\end{minipage}
\caption{\label{fig:eigenmap}
Flattening the Swiss rolls: dimensionality reduction using unimodal (left, center) and multimodal (right) eigenmaps. Joint eigenvectors were computed using JADE.  }
\end{figure}

\begin{figure}[t!]
\centering
\includegraphics[width=1\linewidth]{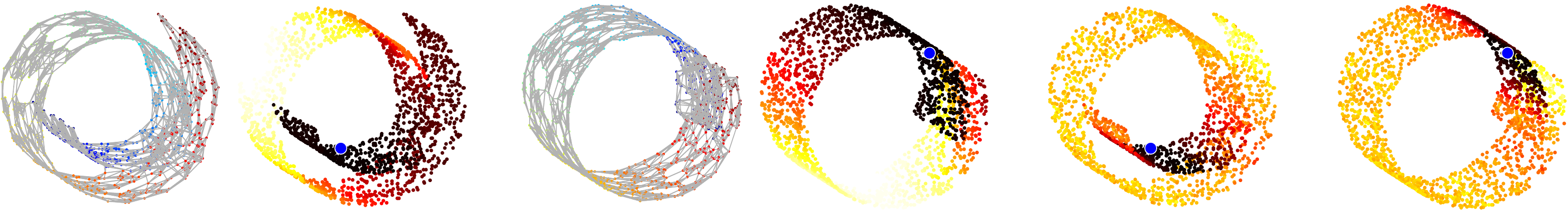} \vspace{1mm}
\begin{minipage}[b]{0.31\linewidth}
\centering \small Modality 1
\end{minipage}
\begin{minipage}[b]{0.31\linewidth}
\centering \small Modality 2
\end{minipage}\hspace{1mm}
\begin{minipage}[b]{0.35\linewidth}
\centering \small Multimodal
\end{minipage}\vspace{-3mm}
\caption{\label{fig:dist}
Diffusion distances from the blue point to the rest of the points on the Swiss roll surfaces. Darker colors represent smaller distances. 
First and third columns show the connectivity used in the construction of the Laplacians. Joint eigenvectors were computed using JADE. 
%
}
\end{figure}

\section{Results}

We tested the proposed approach on three applications: dimensionality reduction, diffusion distance, and spectral clustering. 
All the datasets and code generating the results in this section are available from {\tt anonymous.com}. 
Additional results are shown in the supplementary material. 

{\bf Swiss rolls. }
In the first experiment, we used two Swiss roll surfaces with slightly different embedding as two different data modalities. The rolls were constructed in such a way that in each modality there is topological noise (connectivity ``across'' the roll loops) at different points. Laplacians were constructed as in \cite{Belkin02laplacianeigenmaps} using $5$-neighbor
connectivity and Gaussian weights with scale parameter $t$. 
Figure~\ref{fig:evecs} shows the first few eigenvectors computed using each Laplacian individually and jointly. 
%
%
Figure~\ref{fig:eigenmap} shows two-dimensional embeddings of the same surfaces using the first non-trivial eigenvectors. 
%
When using joint eigenvectors, we are able to correctly capture the intrinsic structure of the data. 
%
%
Figure~\ref{fig:dist} shows the diffusion distance on the Swiss roll surfaces, computed using the first 100 eigenvectors and heat diffusion kernel $K(\lambda) = e^{-1000\lambda}$. Topological noise is clearly visible especially in the first modality, resulting in the distance between two loops to be small. This phenomenon does not occur when using joint eigenvectors.

{\bf Synthetic data clustering. } In the second experiment, we performed clustering on several synthetic multimodal datasets. 
Laplacians were constructed using 15 nearest neighbors (10 for the {\em circles}), and Gaussian weight selected using the self-tuning approach of \cite{perona2004self}.  
%
We compare spectral clustering based on single modalities (SC-1 and SC-2) and joint diagonalization obtained using the JADE method of \cite{Cardoso96jacobiangles}; harmonic mean (JD-HM) of Laplacians (\cite{Cai2011}); and a non-spectral Comraf clustering algorithm (\cite{BekkermanJ07}). Quality was measured using the clustering accuracy criterion as defined in \cite{BekkermanJ07}. For {\em Blobs}, accuracy is averaged over 100 experiments ran on randomly generated datasets.

\begin{figure}[ht]

\begin{minipage}[b]{0.23\linewidth}
\centering
\includegraphics[width=0.9\linewidth]{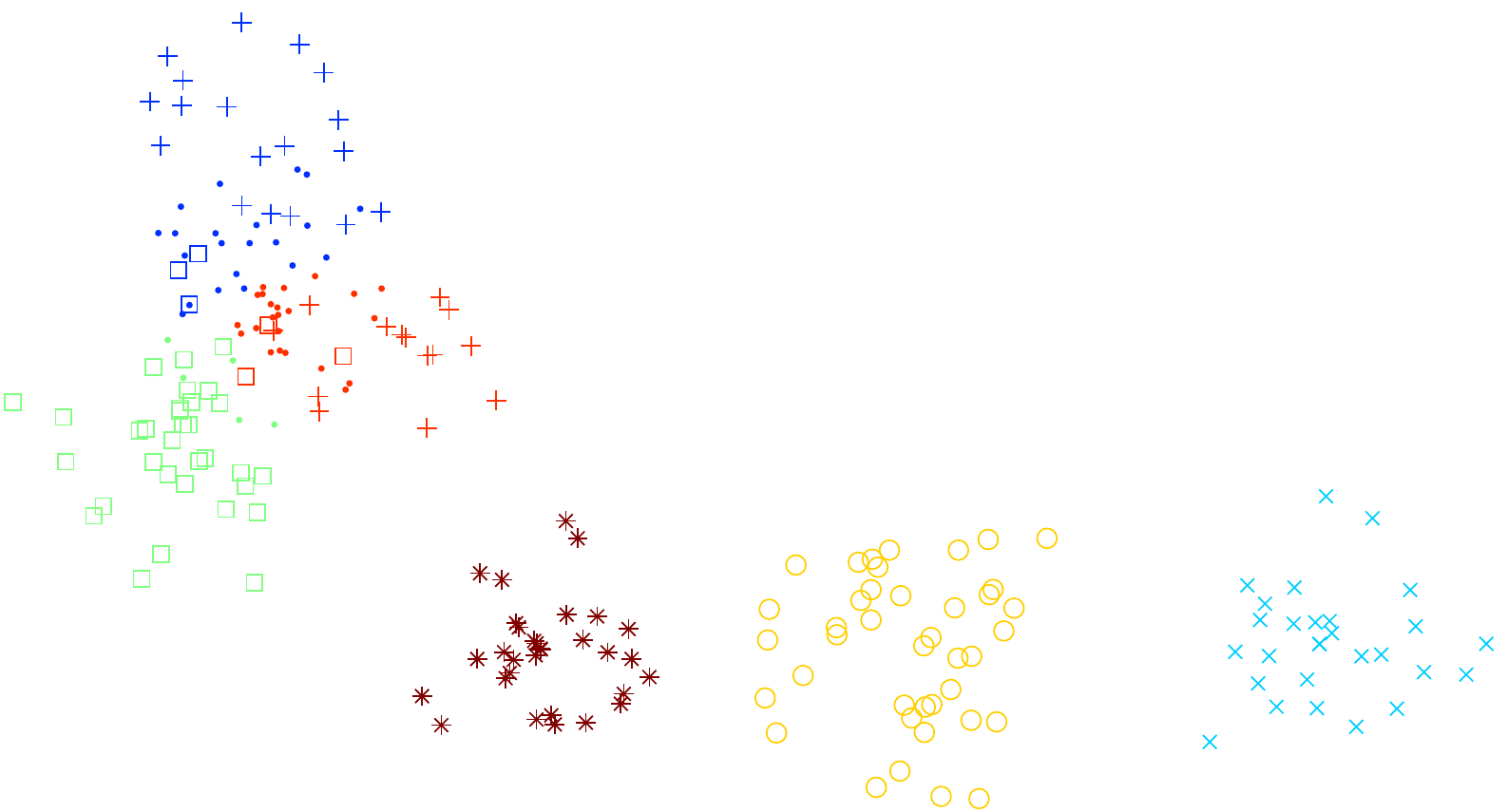} 
\end{minipage}\hspace{1mm}
\begin{minipage}[b]{0.23\linewidth}
\centering
\includegraphics[width=0.9\linewidth]{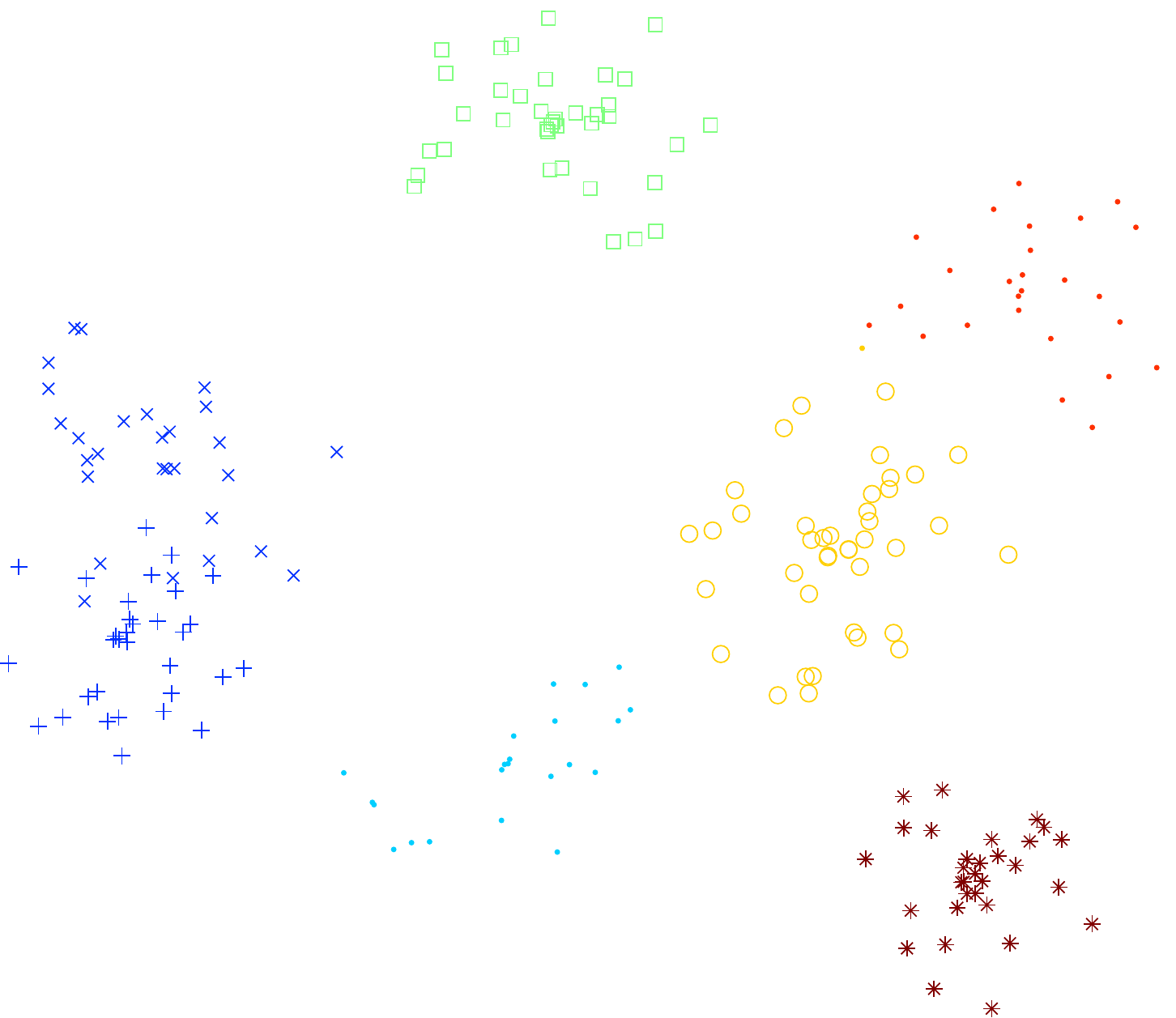} 
\end{minipage}\hspace{6mm}
\begin{minipage}[b]{0.23\linewidth}
\centering
\includegraphics[width=0.9\linewidth]{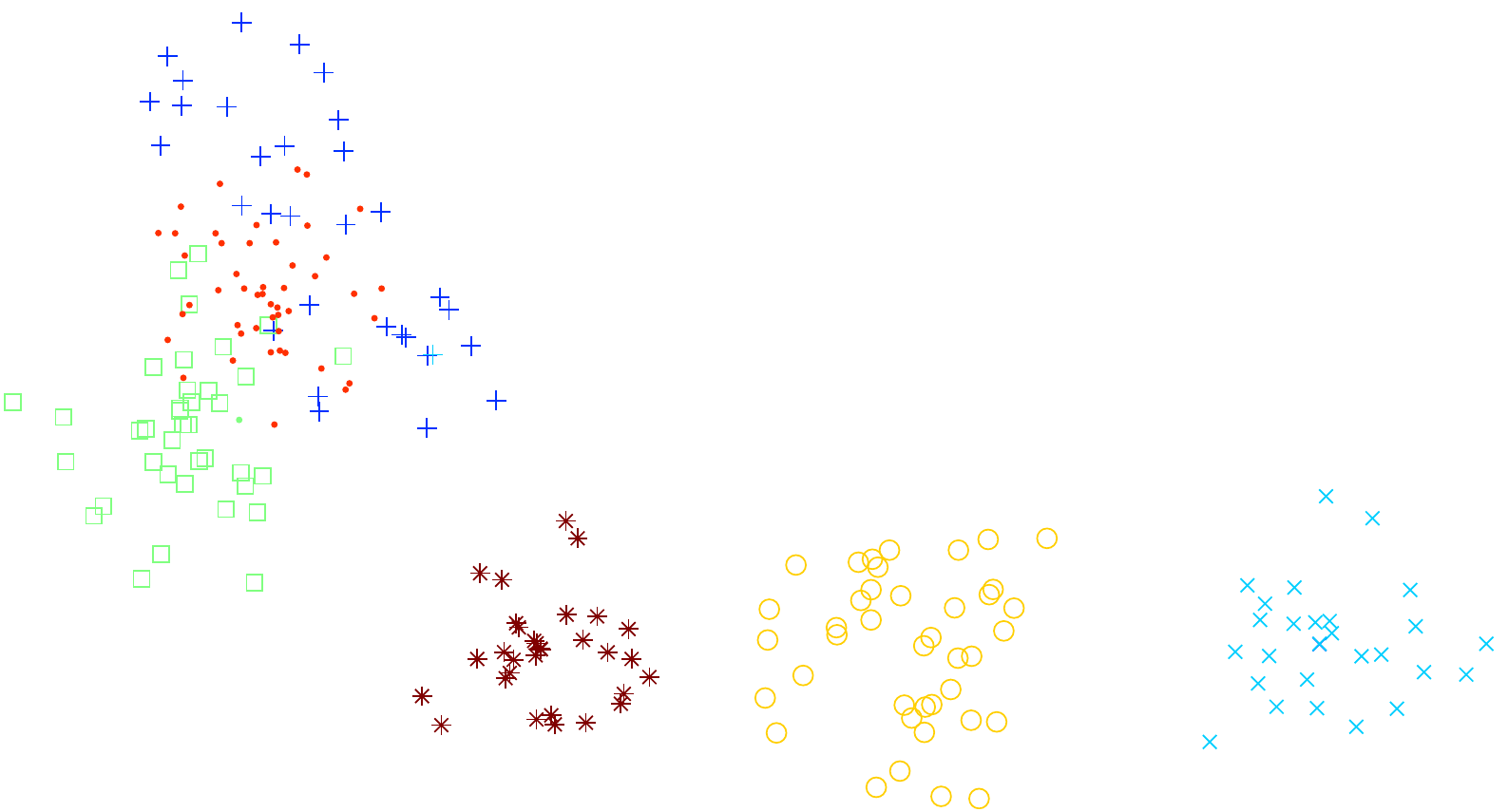} 
\end{minipage}\hspace{1mm}
\begin{minipage}[b]{0.23\linewidth}
\centering
\includegraphics[width=0.9\linewidth]{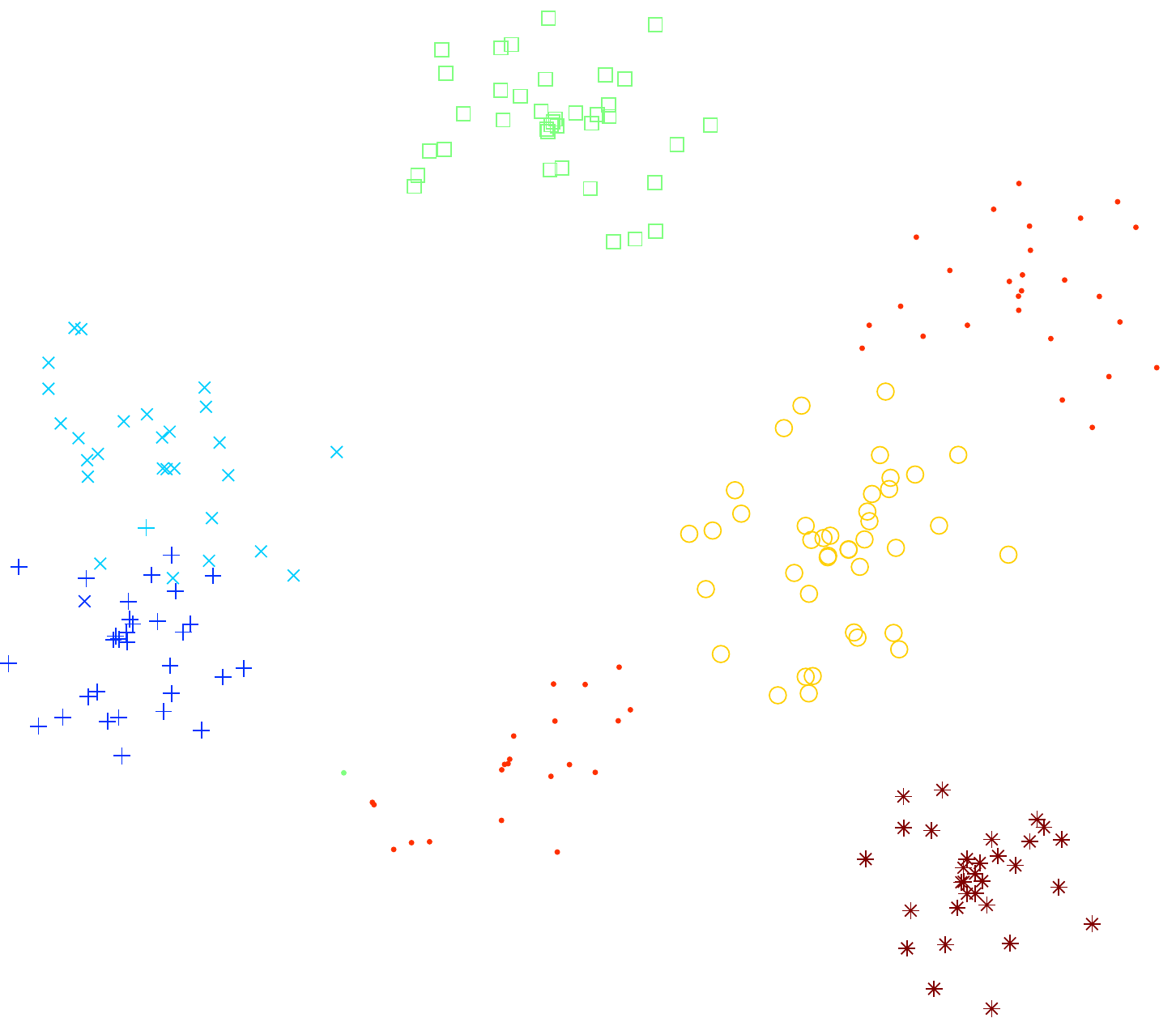} 
\end{minipage}\vspace{2mm}
\begin{minipage}[b]{0.23\linewidth}
\centering
\includegraphics[width=0.8\linewidth]{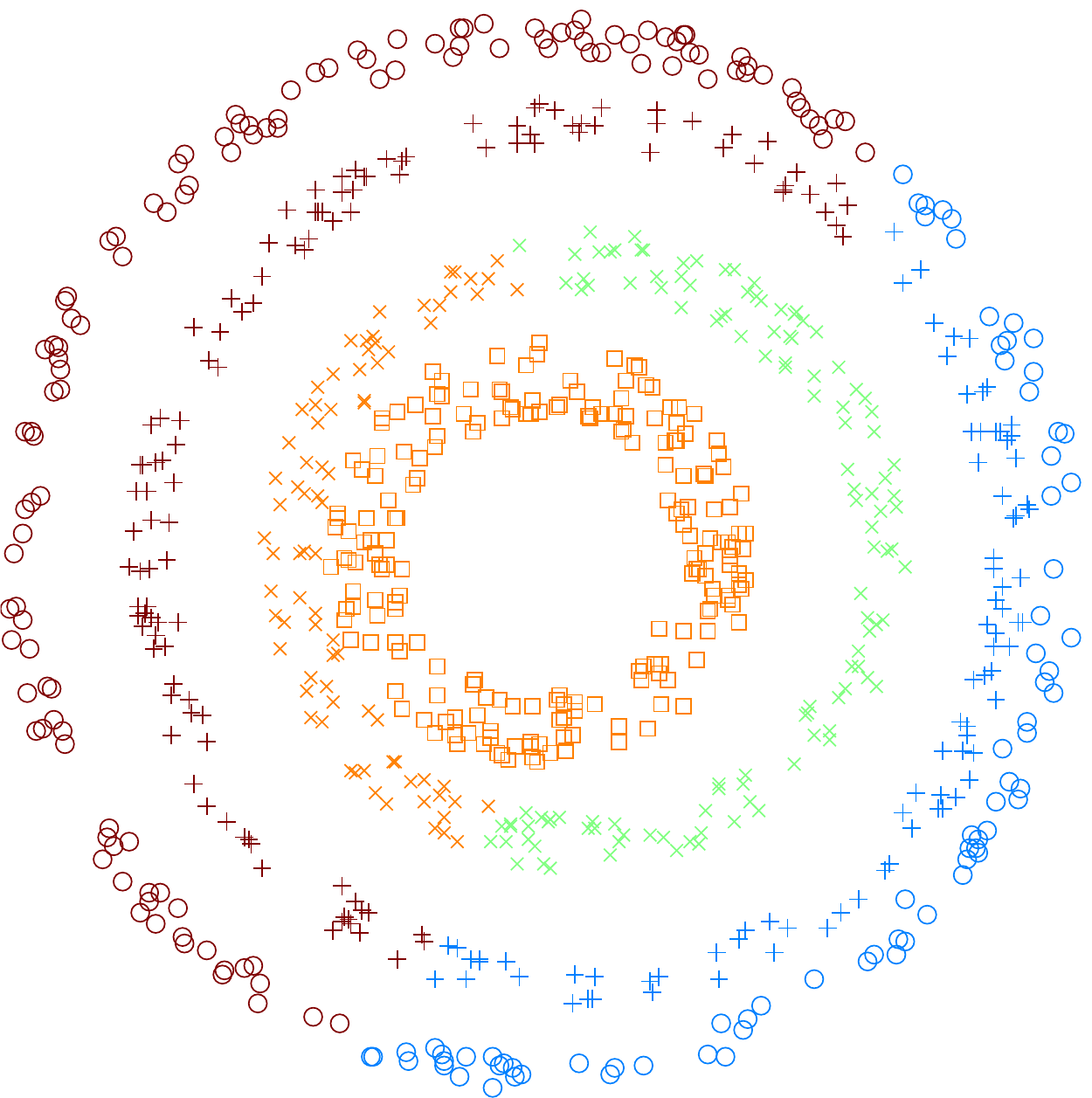} 
\end{minipage}\hspace{1mm}
\begin{minipage}[b]{0.23\linewidth}
\centering
\includegraphics[width=0.8\linewidth]{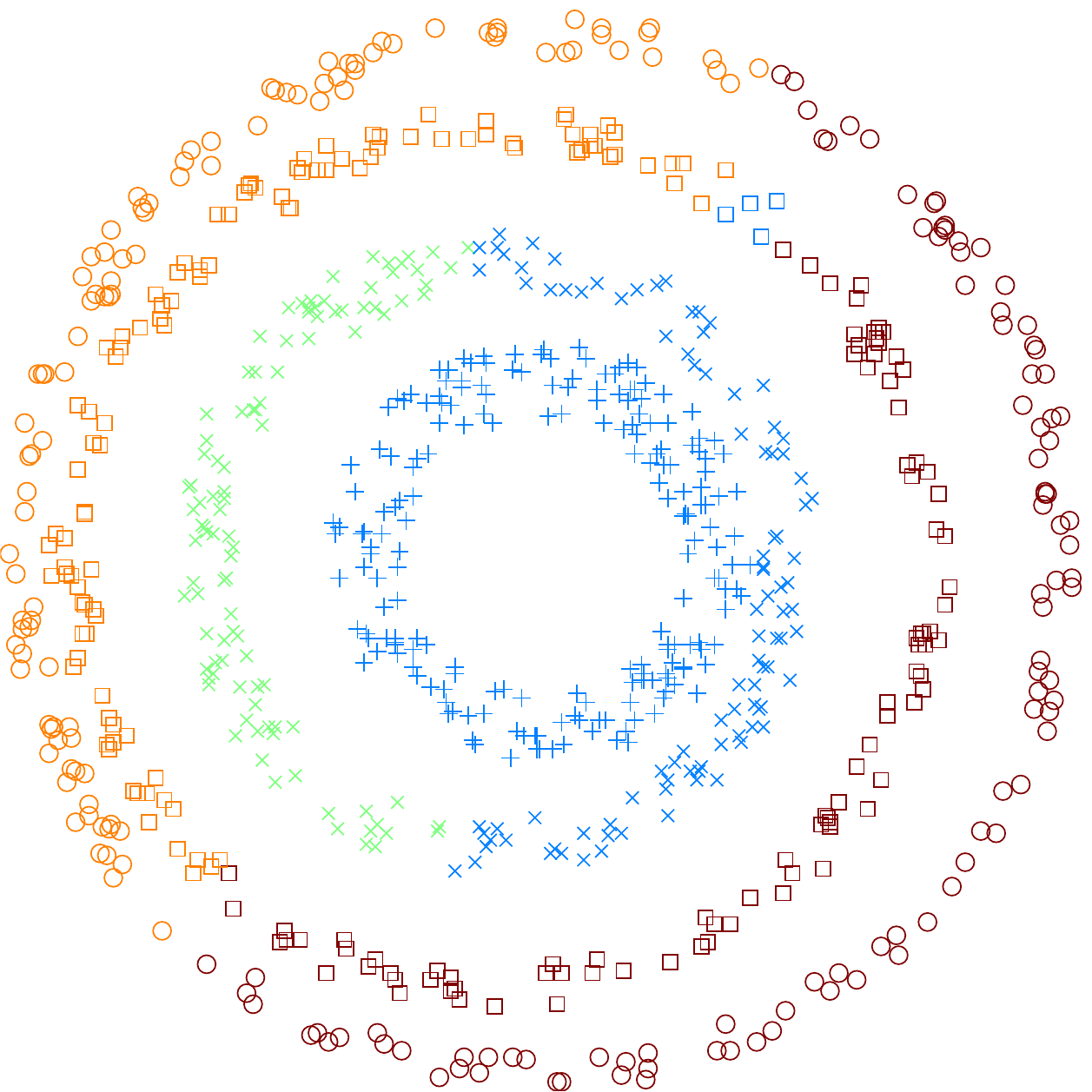} 
\end{minipage}\hspace{6mm}
\begin{minipage}[b]{0.23\linewidth}
\centering
\includegraphics[width=0.8\linewidth]{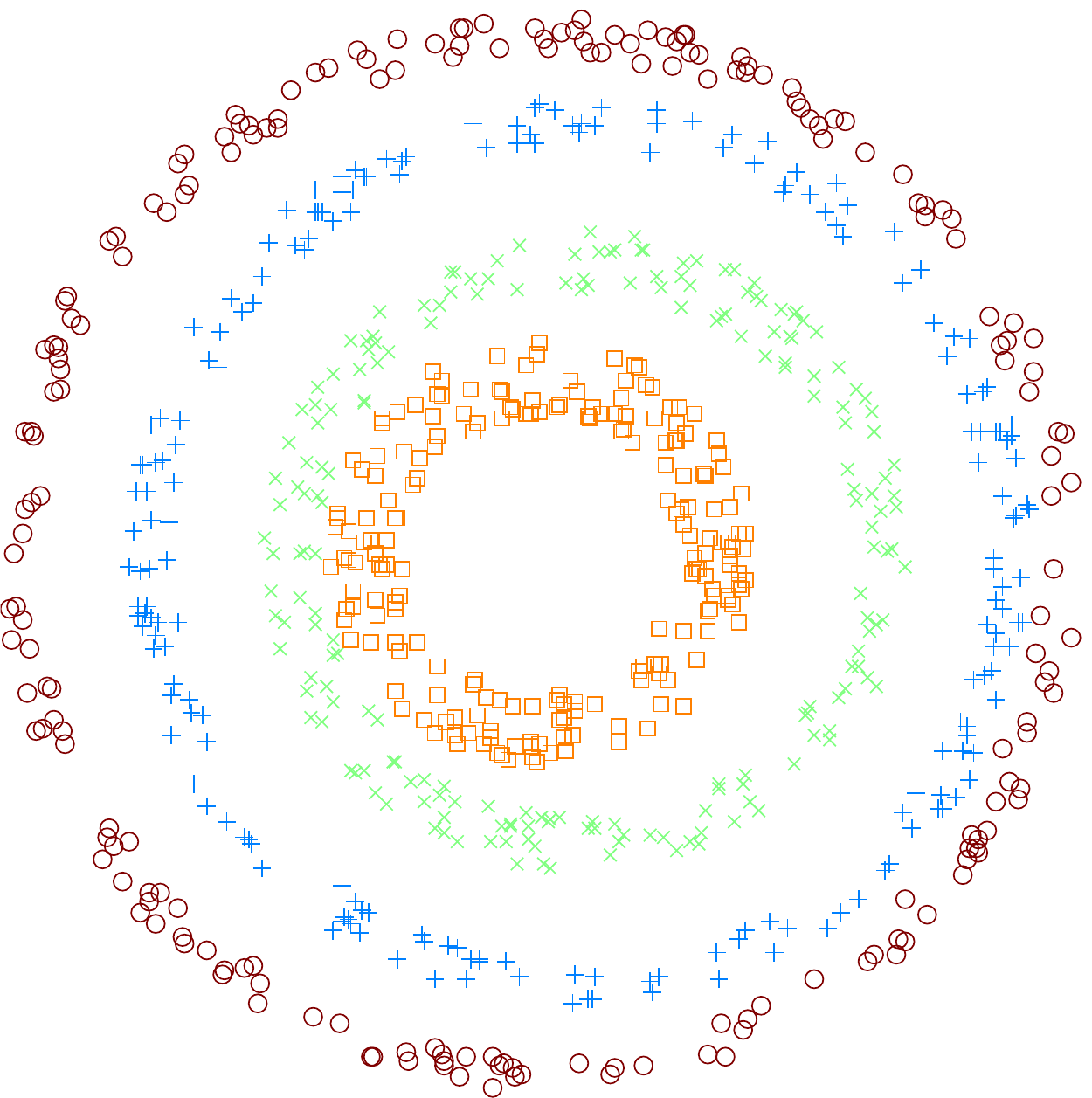} 
\end{minipage}\hspace{1mm}
\begin{minipage}[b]{0.23\linewidth}
\centering
\includegraphics[width=0.8\linewidth]{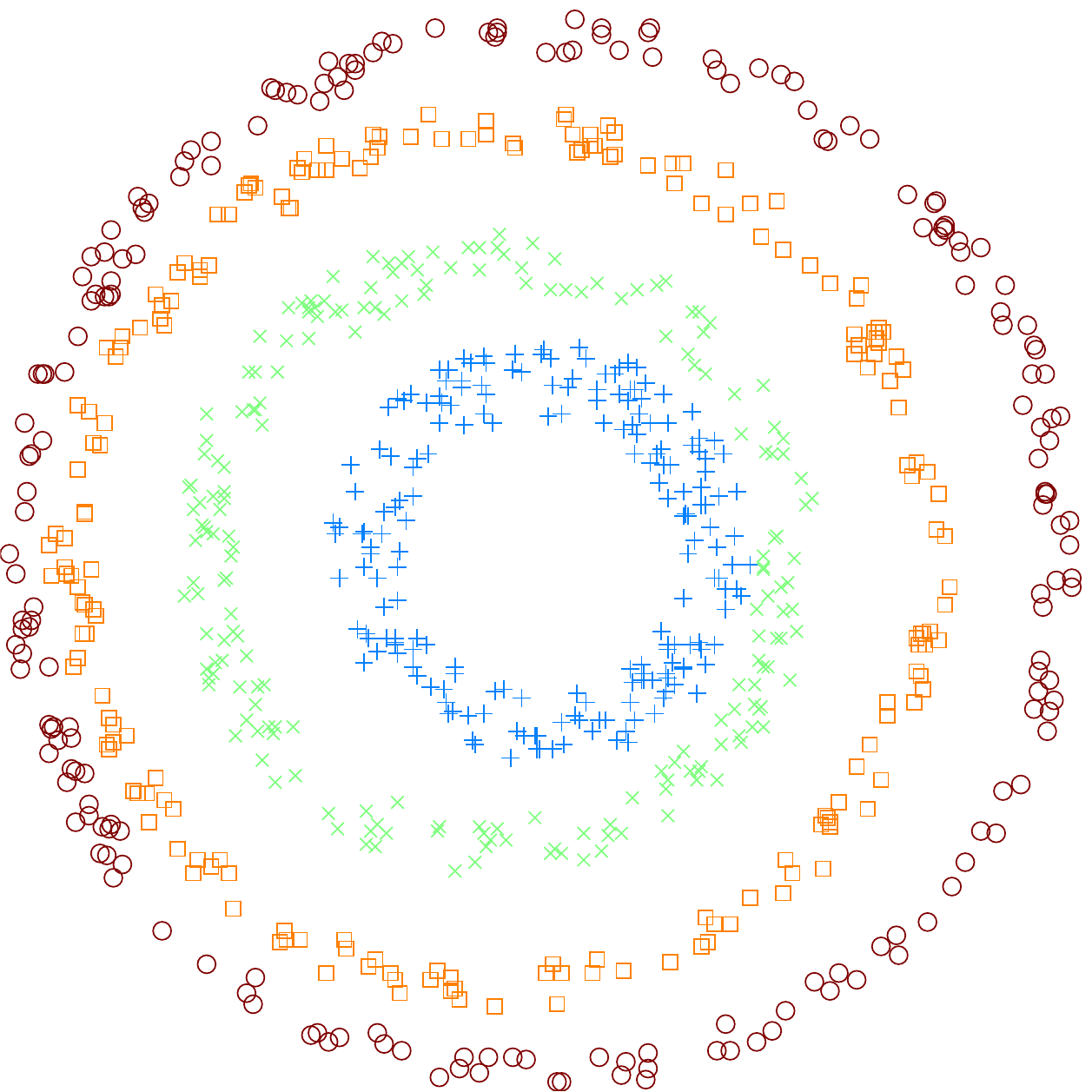} 
\end{minipage}\vspace{1mm}
\begin{minipage}[b]{0.23\linewidth}
\centering
\includegraphics[width=0.9\linewidth]{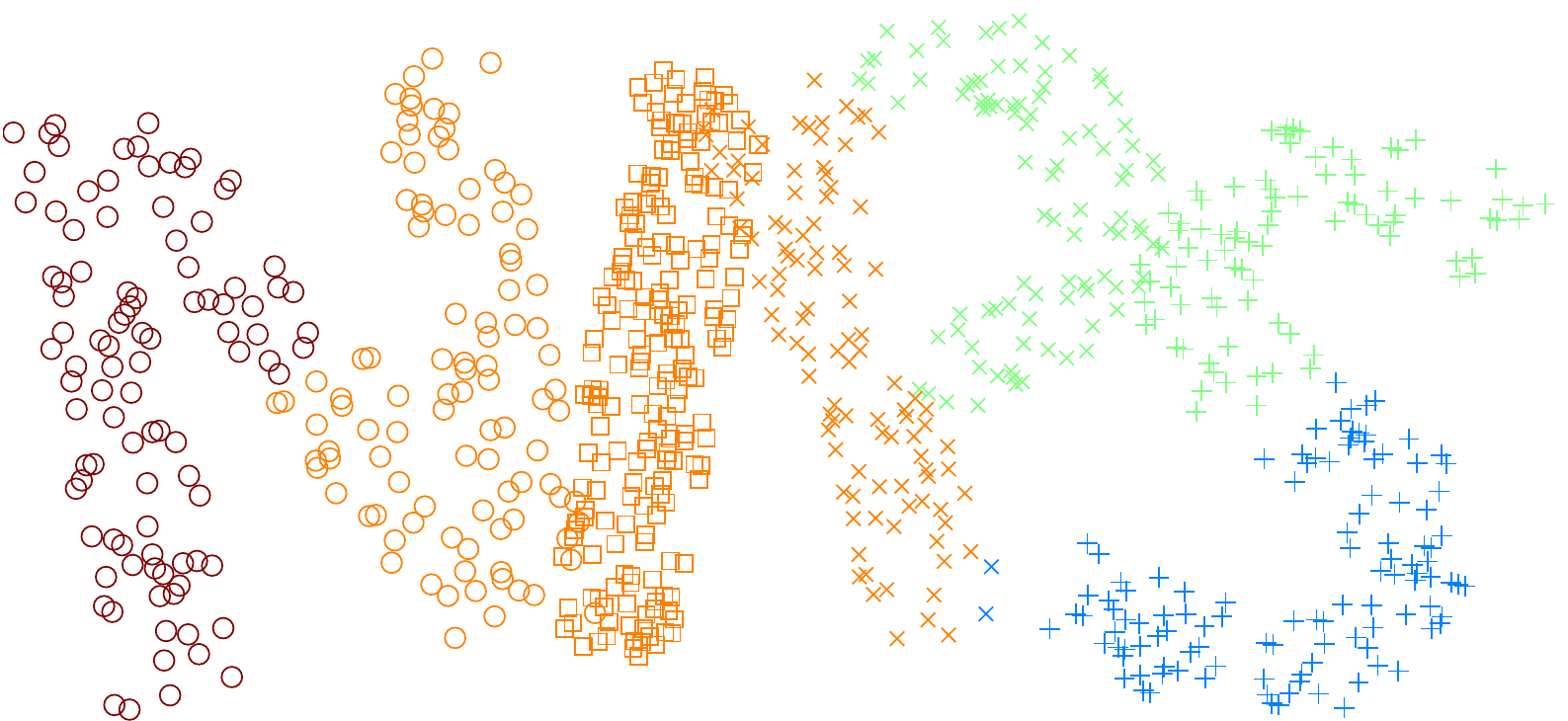} 
\end{minipage}\hspace{1mm}
\begin{minipage}[b]{0.23\linewidth}
\centering
\includegraphics[width=0.9\linewidth]{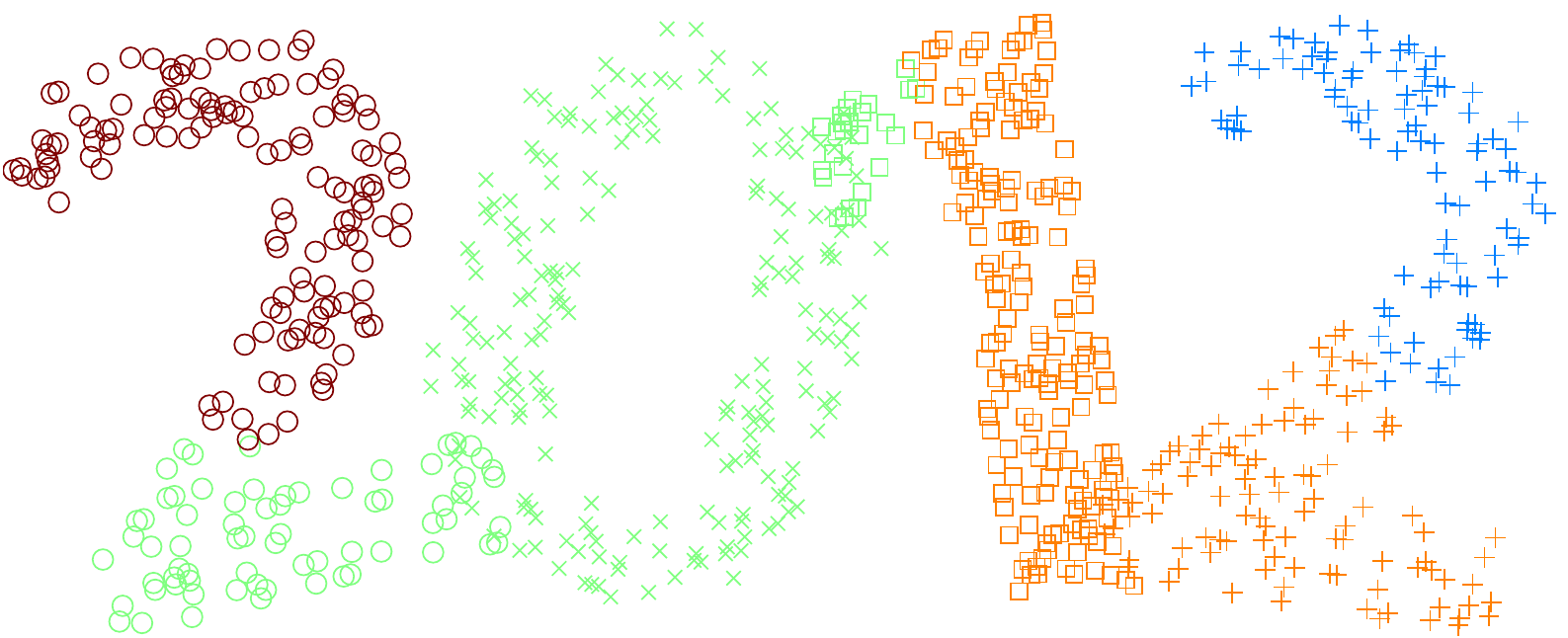} 
\end{minipage}\hspace{6mm}
\begin{minipage}[b]{0.23\linewidth}
\centering
\includegraphics[width=0.9\linewidth]{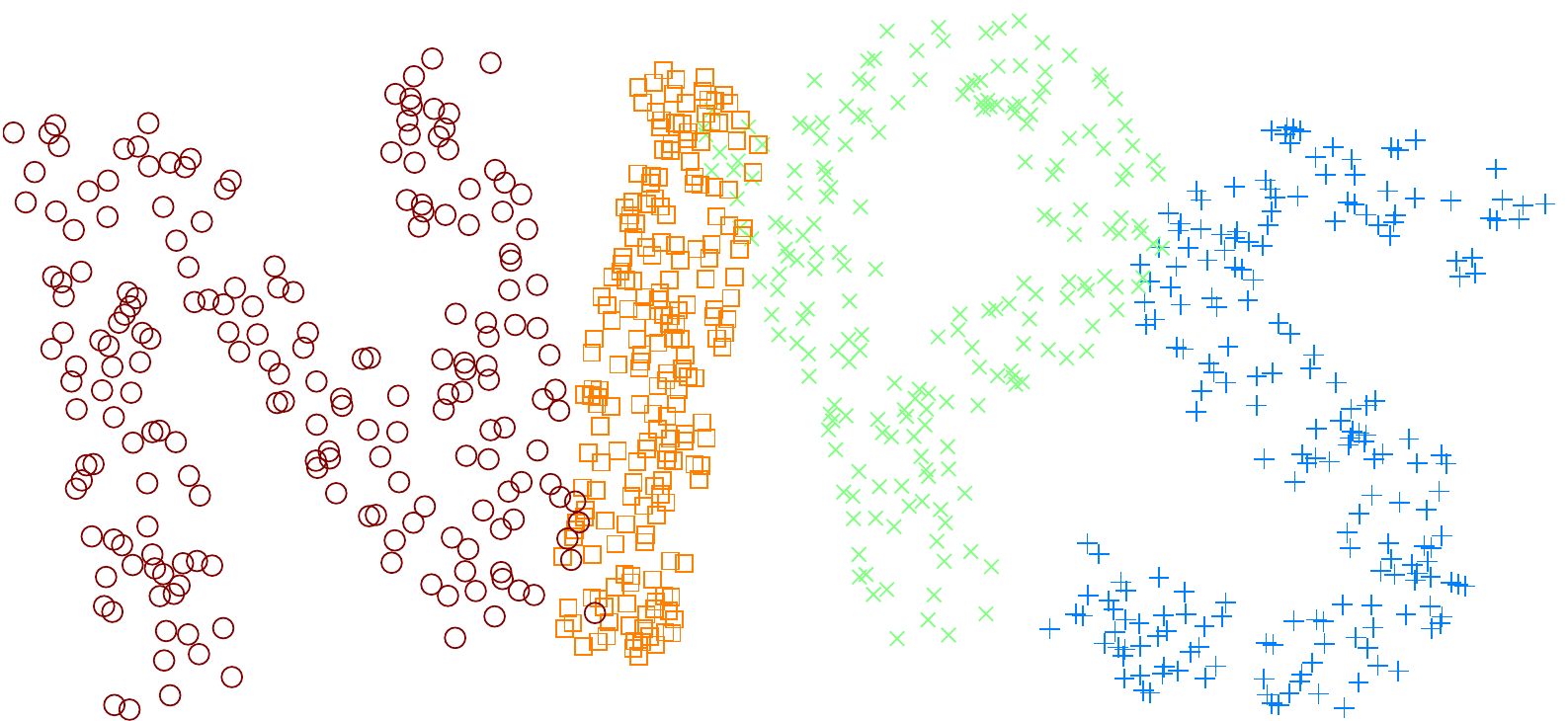} 
\end{minipage}\hspace{1mm}
\begin{minipage}[b]{0.23\linewidth}
\centering
\includegraphics[width=0.9\linewidth]{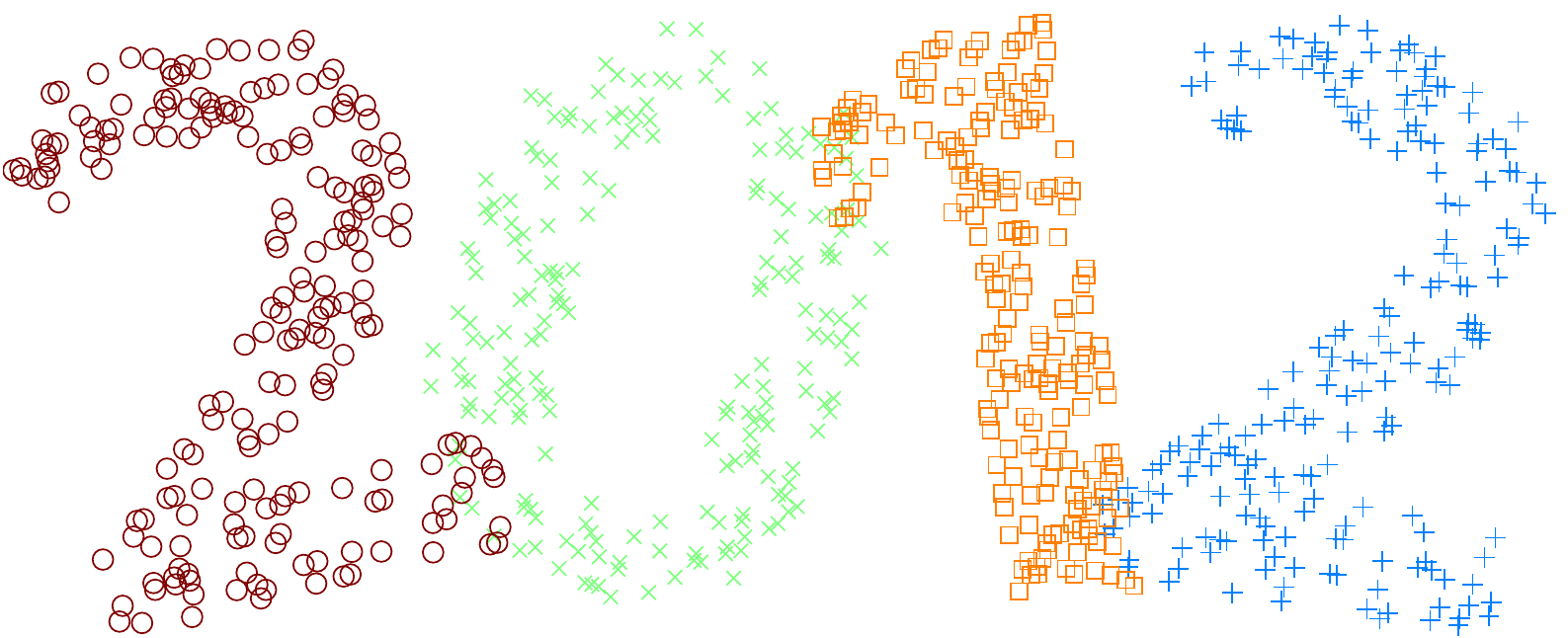} 
\end{minipage}\vspace{-1mm}
\begin{minipage}[b]{0.23\linewidth}
\centering
\small Modality 1
\end{minipage}\hspace{3mm}
\begin{minipage}[b]{0.23\linewidth}
\centering
\small Modality 2
\end{minipage}\hspace{1mm}
\begin{minipage}[b]{0.46\linewidth}
\centering
\small Joint
\end{minipage}\vspace{-3mm}
\caption{\label{fig:clustering1} Clustering synthetics datasets. 
Marker size represents ground truth; marker color represents segmentation results (ideally, markers of each type should have a single color).}
\end{figure}

%
%
The results are summarized in Figure~\ref{fig:clustering1} and Table~\ref{tab:clustering1}.   
%
%
%
Surprisingly, the simple-minded averaging approach performs extremely well; this is consistent with the previously reported results and the success of the methods of \cite{Cai2011} (essentially harmonic mean) and \cite{MaAnchor2008} (arithmetic mean).


\begin{table}[htdp]
\begin{center}
\begin{tabular}{rccccccc}
                        & \small{\bf Clus.} & \small{\bf SC-1} & \small{\bf SC-2} & \small{\bf JADE} & \small{\bf JD-HM} & \small{\bf Comraf} \\
\hline
\small{\em Blobs}	& \small6 & \small91.0$\pm$7.2\% & \small90.8$\pm$7.2\% & \small97.3$\pm$4.2\% & \small98.3$\pm$3.0\% & \small86.9$\pm$8.6\%   \\ 
\small{\em Circles}	& \small4 & \small65.9\% & \small63.4\% & \small{100.0\%} & \small99.8\% & \small31.4\% \\
\small{\em NIPS}	& \small4 & \small63.3\% & \small75.1\% & \small99.9\% & \small99.9\% & \small51.8\% \\
\hline
\small{\em NUS}	& \small7 & \small83.5\% & \small71.0\% & \small92.4\% & \small80.7\% & \small82.1\%  \\
\hline
 \multirow{2}{*}{\small{\em Caltech}}     
	& \small7 & \small73.3\% & \small76.2\% & \small86.7\% & \small84.8\% &  -- \\
	& \small20 & \small66.3\% & \small70.7\% & \small73.3\% & \small76.0\% & -- \\
\hline
\end{tabular}
\end{center}\vspace{-3mm}
\caption{\label{tab:clustering1} Accuracy of different clustering methods.}
\end{table}



\begin{figure}[t!]
\centering
\includegraphics[width=1\linewidth]{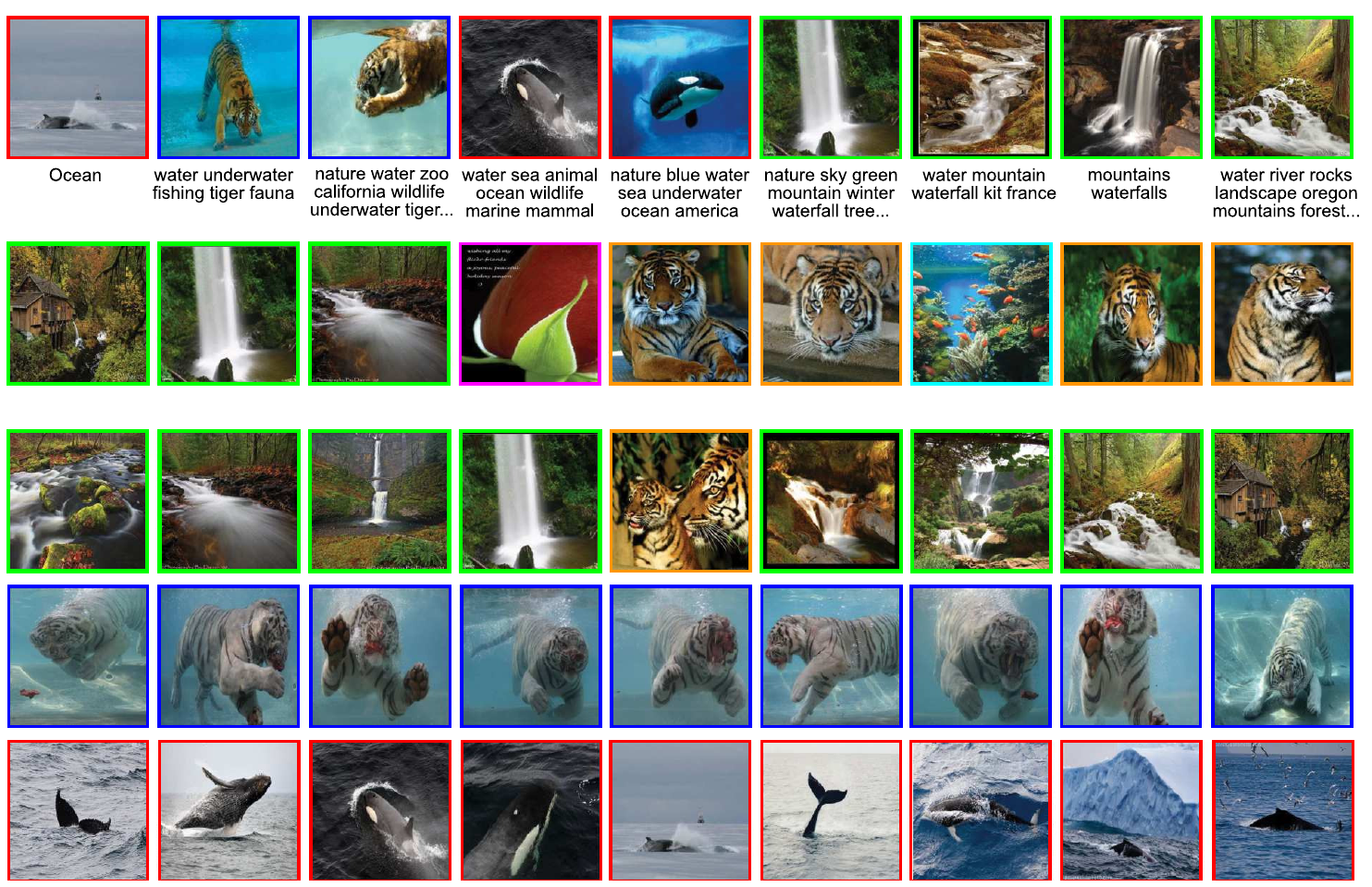} \vspace{-7mm}
\caption{\label{fig:nus}
Spectral clustering of NUS dataset. Shown are a few images and corresponding tags belonging to the same cluster obtained using Tags (top row), Color histogram (second row), and joint modalities (third to fifth row). Groundtruth clusters are shown in different colors. }
\end{figure}

{\bf NUS dataset. }
In the third experiment, we used a subset of the NUS-WIDE dataset \cite{nus-wide-civr09} containing annotated images. The images were selected on purpose to have ambiguous content and annotations (e.g., swimming tigers are also tagged as ``water'' making them confuse e.g. with whales). 
%
As two different modalities, we used the 64-dimensional color histograms and 1000-dimensional bags of words. 
%
Laplacians were constructed using 10 nearest neighbors and Gaussian weight was selected using self-tuning.
Table~\ref{tab:clustering1} shows the performance of different clustering methods, and Figure~\ref{fig:nus} exemplifies the clustered images.

\begin{figure}[t!]
\centering
\includegraphics[width=1\linewidth]{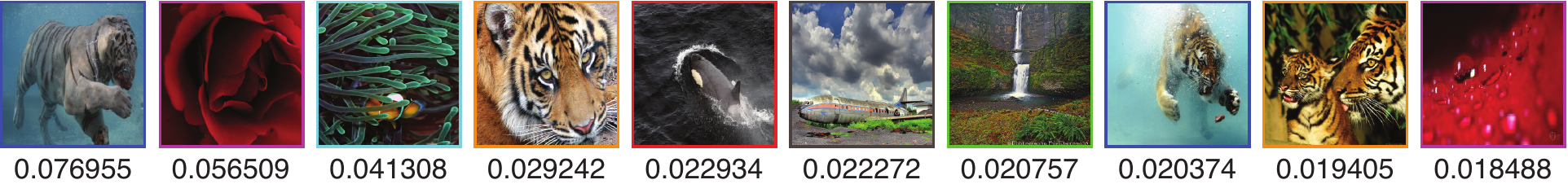}\vspace{1mm} \\
\includegraphics[width=1\linewidth]{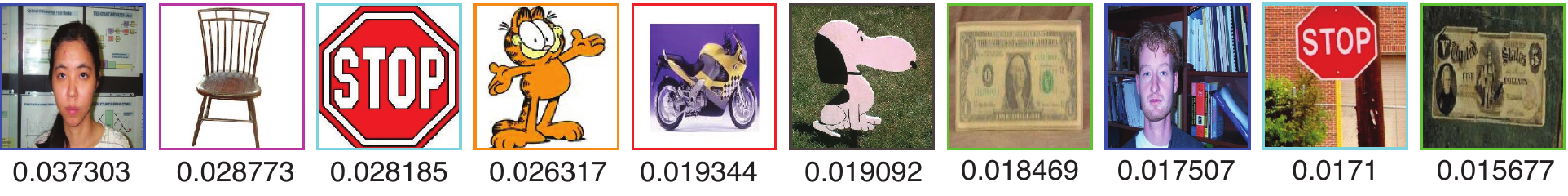} \vspace{-7mm}
\caption{\label{fig:fps}
Farthest point sampling of NUS (top) and Caltech (bottom) datasets using joint diffusion distance.  First point is on the left. Numbers indicate the sampling radius. Note that in both cases, the first 7 samples cover all the image classes. }
\end{figure}

\begin{figure}[t!]
\centering
\includegraphics[width=1\linewidth]{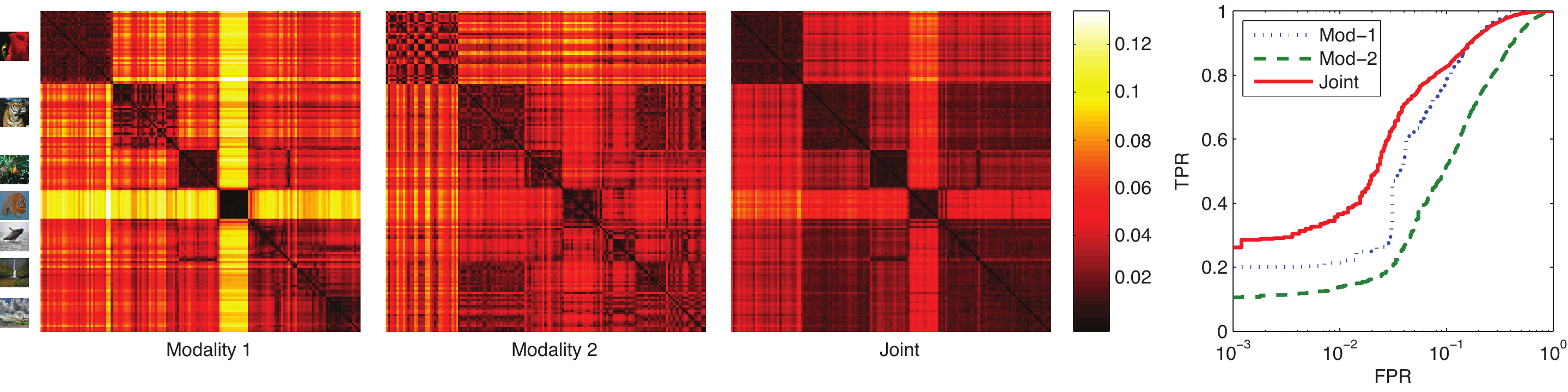}\vspace{0mm}\\
\includegraphics[width=1\linewidth]{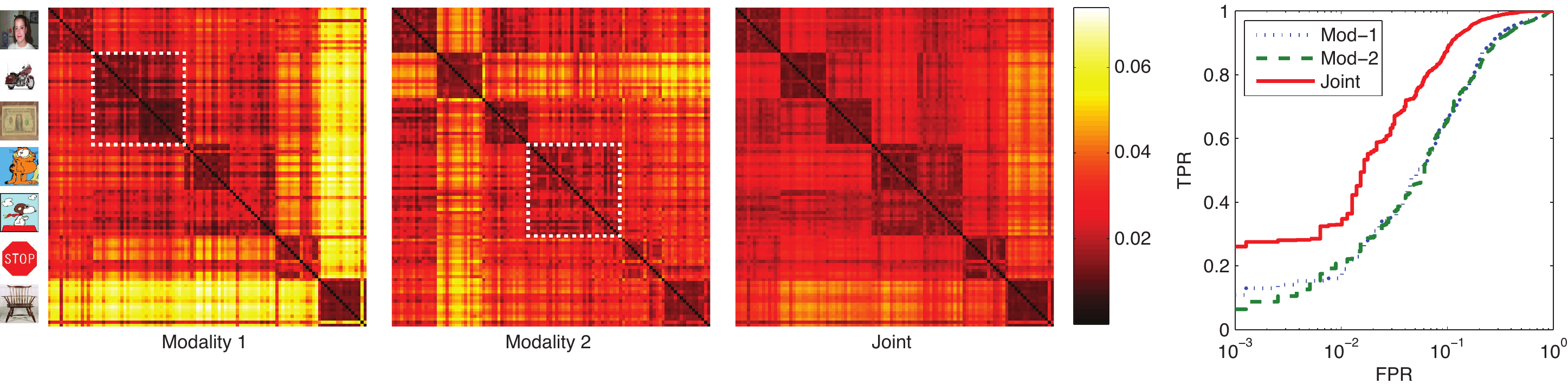} \vspace{-8mm}
\caption{\label{fig:roc}
Columns one to three: distance matrices, column four: ROC curves, computed on NUS (top) and Caltech (bottom) datasets using joint diffusion distance.  Ambiguities are shown in white. }
\end{figure}

Using JADE joint diagonalization, we produced all the joint eigenvectors of the two modalities Laplacians. 
Figure~\ref{fig:roc} (top) shows the distance matrices between the objects in the NUS dataset obtained using uni- and multi-modal diffusion distances (computed with the first $100$ eigenvectors according to~(\ref{eq:diffdist}) using heat diffusion kernel $K(\lambda) = e^{-5\lambda}$). 
Ideally, the distance matrix should contain zero blocks on the diagonal (objects of the same class) and non-zero elsewhere (objects from different classes). Thresholding these distances at a set of levels and measuring the false positives/true positive rates (FPR/TPR), we produce the ROC curves that clearly indicate the advantage of using multiple modalities (Figure~\ref{fig:roc}). 
In Figure~\ref{fig:fps} (top), we used the diffusion distance to progressively sample the NUS dataset using the farthest point sampling strategy: starting with some point, pick up the second one as most distant from the first; then the third as the most distant from the first and second, and so on. Such sampling is almost-optimal (\cite{hochbaum1985best}) and is known to produce a progressively refined $r$-covering of the set.
In fact, the first $7$ samples produced in this way cover all the classes present in the dataset, which is an indication of the meaningfulness of such a sampling.

\begin{figure}[t!]
\centering
\includegraphics[width=1\linewidth]{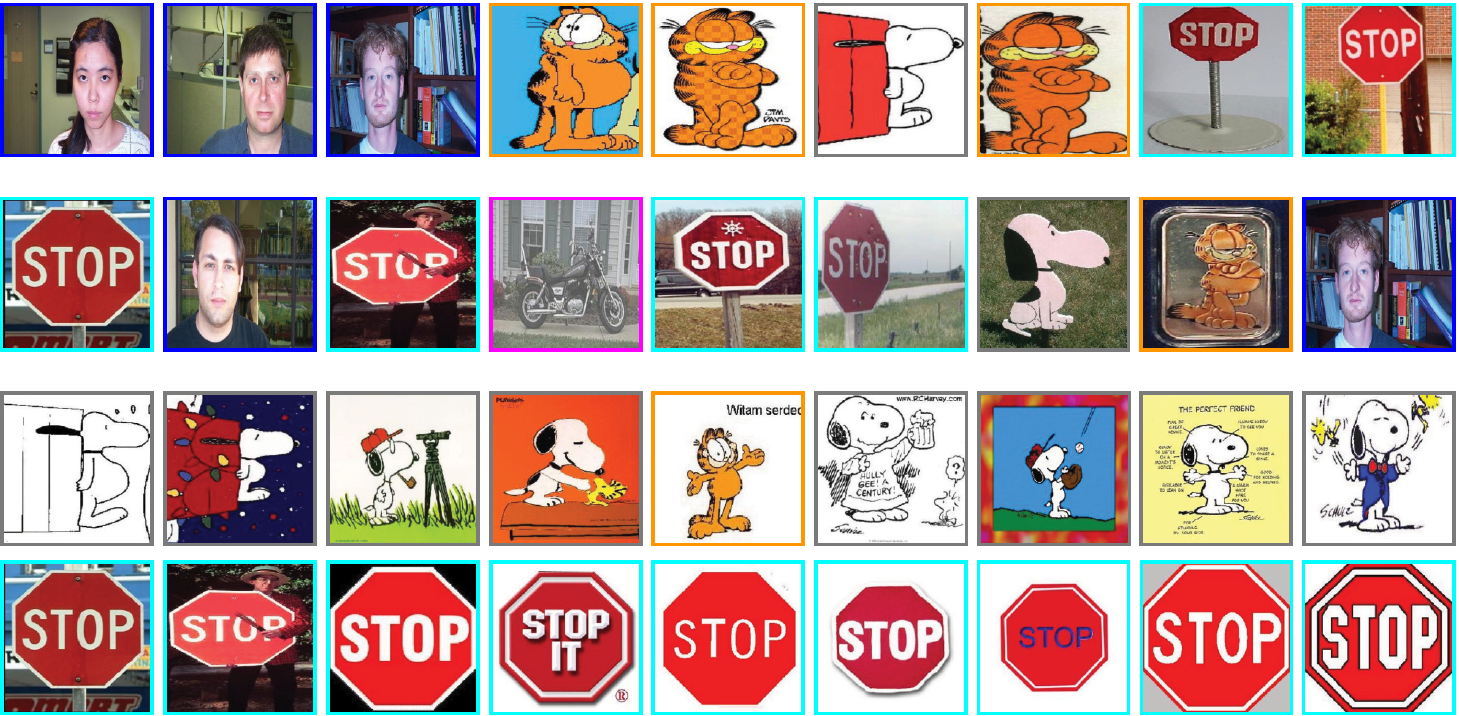} \vspace{-7mm}
\caption{\label{fig:cal}
Spectral clustering of Caltech101 dataset. Shown are a few images and corresponding tags belonging to the same cluster obtained using ht\_bio\_105034 bio-inspired features (top row), 4x4 PHOW (second row), and joint modalities (third and fourth row). Groundtruth clusters are shown in different colors.  }
\end{figure}

{\bf Caltech dataset. }
In the fourth experiment, we repeated the third experiment on a subset of the Caltech-101 dataset with 7 and 20 image classes as in \cite{Cai2011}. 
%
%
For each image, kernels arising from different visual descriptors were given. 
For the 7-clusters experiment, we used the bio-inspired features and 4x4 pyramid histogram of visual words (PHOW); for the 20-clusters experiment, we used geometric blur and 4x4 PHOW descriptors as different modalities, respectively.
%
%
Laplacians were constructed from these kernels using 
Gaussian weight selected with self-tuning.
Diffusion distances were computed with the first $100$ eigenvectors using the kernel $K(\lambda) = e^{-5\lambda}$. 
The results are shown in Figures~\ref{fig:cal}--\ref{fig:roc}.

\section{Discussion and Conclusions}

We presented a framework for multi-modal data analysis using approximate joint diagonalization of Laplacian matrices, naturally extending the 
classical construction of diffusion geometry to the multi-modal setting. This construction allowed an almost straightforward extension of 
various diffusion-geometric data analysis tools such as spectral clustering and manifold learning based on diffusion maps.
In follow-up studies, we intend to show multi-modal extensions of other related techniques such as spectral hashing.

We also showed that many previously proposed approaches to multi-modal spectral clustering are nearly equivalent
and try to solve some version of the joint approximate diagonalization problem.
From the numerical perspective, existing methods were tailored for computing the null joint eigenvectors that are sought for in clustering problems. 
The underlying optimization problems are poorly suited for broader applications of diffusion geometry such as non-linear dimensionality reduction and manifold learning, where many or all eigenvectors of the Laplacians are of interest. 
While approximate joint  diagonalization methods developed in the signal processing community for source separation problems can address the latter case, they were initially developed for full matrices and do not take advantage of the sparse structure of Laplacians. 

To the best of our knowledge, there currently exists no efficient tool to compute the joint eigenvectors of very large sparse matrices, akin Matlab's \texttt{eigs}. 
We believe that the presented construction makes the need of such a tool central enough to deserve the interest of the entire machine learning community. In future work, we will consider extending standard methods for eigendecomposition of large sparse matrices to the joint diagonalization case.

\clearpage
\small
\setlength{\bibspacing}{0.2\baselineskip}
\bibliography{jdpaper}
\bibliographystyle{icml2012}

\end{document}